\documentclass[times,twocolumn,final]{elsarticle}

\usepackage{medima}
\usepackage{framed,multirow}
\usepackage[export]{adjustbox}
\usepackage{amssymb}
\usepackage{latexsym}

\usepackage{url}
\usepackage{xcolor}
\usepackage{bm}
\usepackage{amsmath}

\usepackage{hyperref}

\definecolor{newcolor}{rgb}{.8,.349,.1}

\journal{Medical Image Analysis}

\begin{document}

\verso{Hyeonwoo Cho \textit{et~al.}}

\begin{frontmatter}

\title{Effective Pseudo-Labeling based on Heatmap for Unsupervised Domain Adaptation in Cell Detection}

\author[1]{Hyeonwoo \snm{Cho}\corref{cor1}}
\cortext[cor1]{Corresponding author: Hyeonwoo Cho }
\ead{hyeonwoo.cho@human.ait.kyushu-u.ac.jp}
\author[1]{Kazuya \snm{Nishimura}}
\author[2]{Kazuhide \snm{Watanabe}}
\author[1]{Ryoma \snm{Bise}}
\ead{bise@human.ait.kyushu-u.ac.jp}

\address[1]{Department of Advanced Information Technology, Kyushu University, Fukuoka, Japan}
\address[2]{RIKEN Center for Integrative Medical Sciences, Yokohama, Japan}

\received{1 May 2013}
\finalform{10 May 2013}
\accepted{13 May 2013}
\availableonline{15 May 2013}

\begin{abstract}
Cell detection is an important task in biomedical research. Recently, deep learning methods have made it possible to improve the performance of cell detection. However, a detection network trained with training data under a specific condition (source domain) may not work well on data under other conditions (target domains), which is called the domain shift problem. In particular, cells are cultured under different conditions depending on the purpose of the research. Characteristics, {\it e.g.}, the shapes and density of the cells, change depending on the conditions, and such changes may cause domain shift problems. Here, we propose an unsupervised domain adaptation method for cell detection using a pseudo-cell-position heatmap, where the cell centroid is at the peak of a Gaussian distribution in the map and selective pseudo-labeling. In the prediction result for the target domain, even if the peak location is correct, the signal distribution around the peak often has a non-Gaussian shape. The pseudo-cell-position heatmap is thus re-generated using the peak positions in the predicted heatmap to have a clear Gaussian shape. Our method selects confident pseudo-cell-position heatmaps based on uncertainty and curriculum learning. We conducted numerous experiments showing that, compared with the existing methods, our method improved detection performance under different conditions. 
\end{abstract}

\end{frontmatter}


\begin{figure}[t]
    \centering
    \includegraphics[width=\columnwidth]{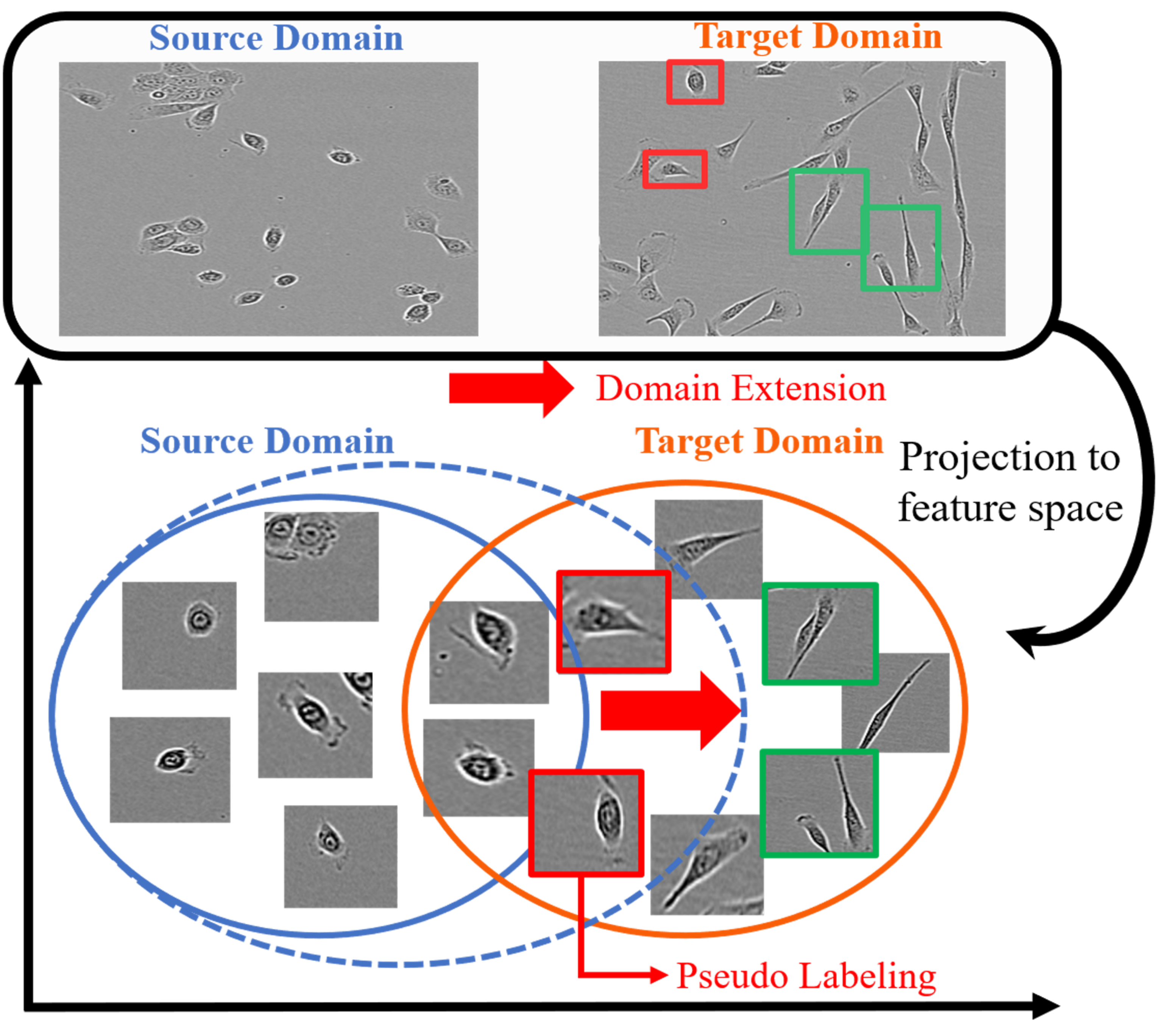}
    \caption{Overview of domain extension. Top shows examples of cell images on source and target domain, and red and green rectangles show cells in target with a similar and different shape to those in source, respectively. Bottom shows an illustration of feature space of patches cut from the top images, where red arrow indicates domain extension.}
    \label{fig:our idea}
\end{figure}

\section{Introduction}
Cell detection, which recognizes individual cell positions, plays a crucial role in various cell analyses, such as cell motility and cell proliferation in the biological processes.
Detecting cell instances manually in huge living cell datasets is too time-consuming for biomedical researchers. Automatic cell detection methods are required for cell analysis. These methods can be classified into conventional image-processing methods and deep-learning methods.

In general, image-processing methods are designed using the image characteristic of the target images. Often these methods only work under the specific conditions used for their development.

Deep-learning methods, such as convolutional neural network (CNN), have recently shown very good performance in cell detection\cite{yi2019multi, li2019signet, kainz2015you, nishimura2019weakly, fujita2020cell}. These methods work well in the same conditions {\it e.g.}, culture condition, as the training data (the source domain). However, they may not work well in other conditions (the target domain) since the characteristics of the images may differ; this is called the domain shift problem. 
In particular, cells grown under different culture conditions have different characteristics {\it e.g.}, different shapes and density, even if the type of cell is the same. Fig. \ref{fig:our idea} shows images of cells grown under different culture conditions. The shapes in the target domain are elongated compared with those in the source domain (the F1-score decreased from 0.941 (source) to 0.768 (target) in our experiments). This indicates that training data needs to be prepared for not only the cell type to be analyzed but also different culture conditions, even though the conditions are very dependent on the purpose of biological research being conducted. Our aim is to address this domain shift problem.

Unsupervised domain adaptation (UDA) methods \cite{ghifary2016deep,ganin2015unsupervised, tzeng2017adversarial, ge2020domain, jin2018unsupervised, saito2019strong, haq2020adversarial, tsai2018domain}, which use only training data in the source domain without training data in the target domain, have been widely used for handling domain shift problems. Most of the previous methods are designed for classification and segmentation tasks, not for detection tasks. A main approach of UDA is adversarial learning, which transfers the distribution of the target to the source's in the same feature \cite{tzeng2017adversarial, ganin2015unsupervised, cui2020gradually, haq2020adversarial}. In particular, Haq {\textit{et al}.} extended this approach to include the cell-segmentation problem by introducing an auto-encoder \cite{haq2020adversarial}. This method tries to transfer the feature distribution of an entire image, in which it is implicitly assumed that the characteristics of the entire image, such as illumination and color,  differ between the source and target domains. However, to reach our goal, this method faces a difficulty regarding domain adaptation of the entire image, as it would contain many cells with various appearances ({\it e.g.}, different shapes and density), as shown in the red and green boxes in Fig. \ref{fig:our idea}. Here, it is very important to take advantage of the different appearances of cells in an entire image.

Another approach to domain adaptation utilizes selective pseudo labels from data in the target domain, where the network is trained using pseudo labels as additional training data \cite{choi2019pseudo, fu2021unsupervised, wang2020unsupervised, long2013transfer,zhang2017joint, pei2018multi}; our method is categorized as this kind of approach. The main idea of pseudo-labeling is that some samples in the target domain can be correctly classified, and if we can correctly select predicted samples as additional training data, the performance in the target domain will improve. 
However, most pseudo-labeling-based methods were designed for classification tasks, and some challenges still remain for detecting densely distributed cells (please refer to the related works in detail).

In this paper, we propose a domain adaptation method for cell detection that works well on different conditions in the same type of cell dataset. 
To handle cells with various appearances in an entire image, we separate the entire image into patches. Then, we assume that some patches in the target are similar to some in the source, {\it i.e.}, the image feature distributions between domains partially overlap like most UDA methods assumption, as shown in Fig. \ref{fig:our idea}.
In our preliminary study, we used a state-of-the-art heatmap-based method for cell detection \cite{cao2019openpose, nishimura2019weakly}, where the cell centroid is at the peak of a Gaussian distribution in the map.
Our key observation is that this method can detect a cell that has a similar but slightly different shape from the source's even when the prediction map shows a distorted (non-Gaussian) shape (see $D(\bm{x}_i^t)$ in Fig. \ref{fig:overview_of_method}).
Furthermore, we found that the detection performance could be improved using the correctly detected cells with clear Gaussian maps (pseudo-cell-position heatmaps) as additional training data.
However, the pseudo labels (pseudo-cell-position heatmaps) may contain many noisy labels that have adverse effects on the detection performance. To select confident pseudo labels, we introduce Bayesian discriminator and curriculum learning. Given the original image and its generated pseudo heatmap, the Bayesian discriminator classifies the heatmap as correct or not with uncertainty. A sample with low uncertainty can be considered as a likely 'correct' heatmap. This heatmap selection contributes to improving detection performance.
We then use the selected patches with clear Gaussian maps as pseudo-cell-position heatmaps for re-training the detector. This process, called domain extension, incrementally adds the confident pseudo labels to improve detection performance in both the source and target domains (Fig. \ref{fig:our idea}). 

In addition, we extend our method proposed in MICCAI2021 \cite{cho2021cell} to select more correct pseudo labels by adding a curriculum learning module to the Bayesian discriminator. Our curriculum learning first selects easier cases and then gradually covers difficult cases step-by-step based on the number of cells in a patch ({\i.e.}, cell density); fewer cells in a patch, easier cell detection. At the beginning of learning, the method selects pseudo heatmaps that contain fewer cells and gradually increases the number with the iteration of learning.
In experiments, we demonstrate the effectiveness of our method for unsupervised domain adaptation using 14 pairs of domains.
The main contributions of this paper are as follows:

\textbf{Contribution:}
\begin{itemize}
  \item We propose an unsupervised domain adaptation method for the cell detection that handles the domain shift problem by using the pseudo-cell-position heatmap that can incrementally extend the domain from the source to the target.
  \item We propose the pseudo-cell-position heatmap as a new pseudo-labeling method for cell detection by using correct peak positions in prediction maps.
  \item To select confident pseudo labels, we introduce a Bayesian discriminator that estimates the uncertainty of each patch under a self-training framework. In addition, we add the curriculum learning module for selecting pseudo labels based on the number of cells to select more confident them step by step.
  \item To confirm the effectiveness of the curriculum learning, we evaluated the proposed method with 14 combinations of different domains on 2 datasets, where one dataset contains 4 conditions (12 pairs), and the other has 2 conditions (2 pairs). We considerably improved our previous results on these experiments. In addition, in the ablation study, we confirmed that the Bayesian discriminator also improved by using curriculum learning.
\end{itemize}
We here note that the extension from the conference version \cite{cho2021cell} contains not only the methodological extension but also a detailed analysis of the original method and extended method, introducing curriculum learning, additional pairs of domains, an analysis of the relationship of the number of cells and the correct pseudo labels, an ablation study of each module, and a detailed explanation of the proposed method.

\section{Related work}
\subsection{Cell detection}
\subsubsection{Cell detection based on image processing}
Many image-processing-based methods have been proposed for cell detection \cite{otsu1979threshold, yuan2012quantitative, cosatto2008grading,chalfoun2014fogbank, al2009improved,biseR2015cell}. They use image features to detect the cells. In particular, Otsu {\textit{et al}.} and Yuan {\textit{et al}.}  utilized a threshold selection from the target images \cite{otsu1979threshold, yuan2012quantitative}, while Cosatto {\textit{et al}.} used 2D image-filters \cite{cosatto2008grading}. Chalfoun {\textit{et al}.} proposed a watershed-based method that uses intensity or edge information \cite{chalfoun2014fogbank}. Al-Kofahi {\textit{et al}.} leveraged the graph-cuts binarization algorithm that extracts connected clusters of nuclei \cite{al2009improved}. However, since these methods are designed for the specific conditions used in their development, they require time to tune the parameters to the specific conditions of the target image.

\subsubsection{Supervised cell detection}
Deep learning-based a convolutional neural network (CNN) has popularly been used for cell detection and localization \cite{yi2019multi, li2019signet, nishimura2019weakly, fujita2020cell}. Given training data on each condition, deep-learning methods outperform image-processing-based methods on datasets having various conditions. He {\textit{et al}.} proposed Mask R-CNN, which enables instance segmentation by adding mask branches to the head of the Faster R-CNN architecture \cite{ren2015faster} and thereby allowing segmentation of each detected object \cite{he2017mask}. Fujita {\textit{et al}.} utilized Mask R-CNN for cell detection and segmentation \cite{fujita2020cell}. To reduce annotation costs, Nishimura {\textit{et al}.} proposed U-Net \cite{ronneberger2015u}, which uses a heatmap as training data \cite{nishimura2019weakly}. However, these methods would not work well at cell detection if the domains of the training data (source domain) and test data (target domain) are different (domain shift). In addition, annotating the image for each condition is very laborious when the individual images contain many cells. 

\subsection{Unsupervised domain adaptation (UDA)}
\subsubsection{Adversarial learning}
Many unsupervised domain adaptation methods have been proposed \cite{ghifary2016deep,ganin2015unsupervised, tzeng2017adversarial, ge2020domain, jin2018unsupervised, saito2019strong, haq2020adversarial, tsai2018domain}. Ganin {\textit{et al}.} and Tzeng {\textit{et al}.} proposed adversarial learning for domain adaptation that transfers the distribution of the target domain to the source domain \cite{ganin2015unsupervised,tzeng2017adversarial}. This method uses a domain discriminator that distinguishes the source and target domain and, for adaptation, it tries to fool the discriminator into not distinguishing between domains. For the cell segmentation task, Haq {\textit{et al}.} utilized adversarial learning for domain adaption and introduced an auto-encoder to extract invariant features between the source and target domains \cite{haq2020adversarial}. However, this approach does not consider essential features {\it e.g.},(class features in classification task) and only tries to match features between domains. To solve this problem, Saito {\textit{et al}.} proposed to use the maximum classifier discrepancy that matches features among classes between domains \cite{saito2018maximum}. 
The design of an adversarial domain adaptation is more complex for the heatmap prediction task since the network has both an encoder and decoder. Adaptation can be applied to the input level, feature level, or output level, and the most suitable level may depend on the dataset.

\begin{figure*}[h]
    \centering
    \includegraphics[width=\linewidth]{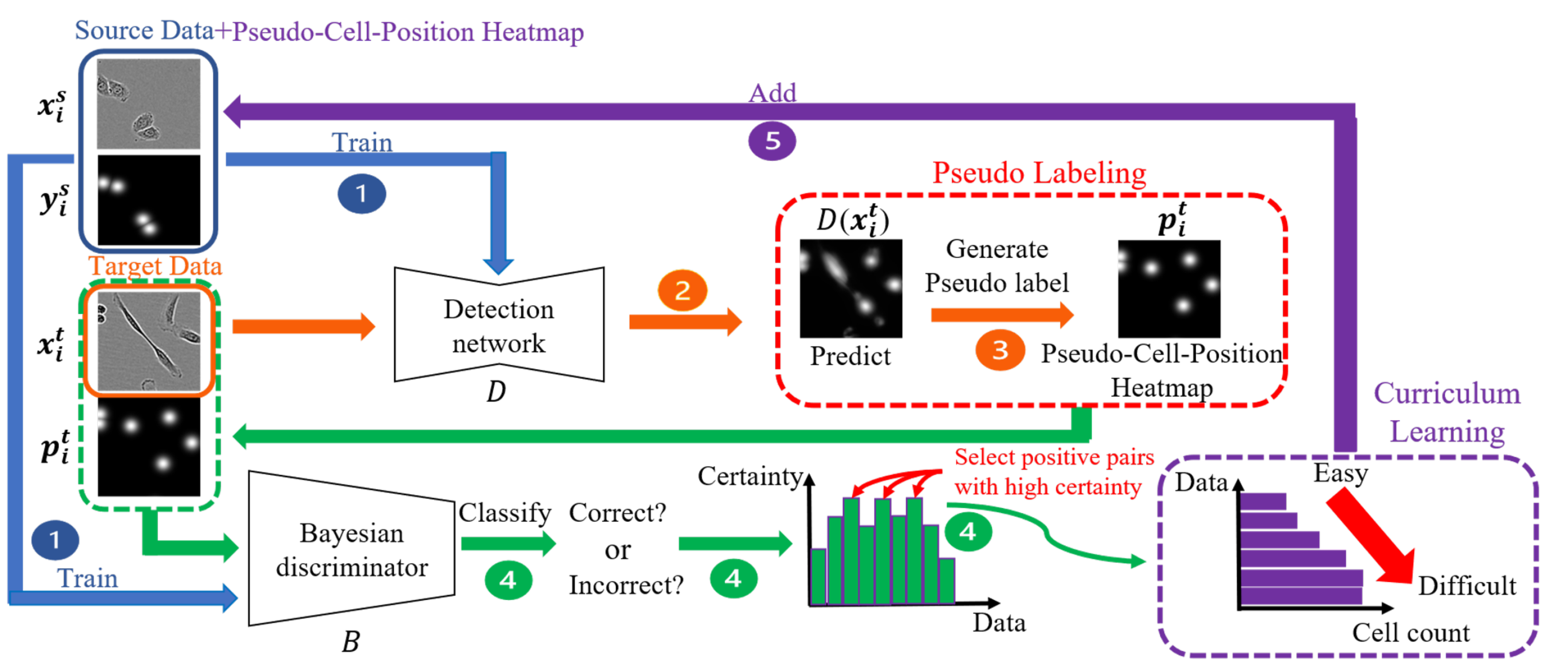}
    \caption{Schematic overview of the proposed method, which consists of five stpdf; Step 1 (blue arrows): training the detection network and Bayesian discriminator with the source data and pseudo labels; Step 2 and 3 (orange): estimating target data and generating candidates of pseudo-cell-position heatmap; Step 4 (green): selecting positive pseudo heatmaps with low uncertainty by Bayesian discriminator; Step 5 (purple): selecting final pseudo heatmaps from the results of the previous step using curriculum learning.}
    \label{fig:overview_of_method}
\end{figure*} 

\subsubsection{Pseudo-labeling}
Pseudo-labeling methods have been used for unsupervised domain adaptation \cite{choi2019pseudo, fu2021unsupervised, wang2020unsupervised, long2013transfer,zhang2017joint, pei2018multi}. 
These methods can be roughly divided into ones that use label selection and ones that do not use label selection.
Pseudo-labeling without selection means that all pseudo labels in the target domain are used as training data. Long {\textit{et al}.} and Zhang {\textit{et al}.} proposed hard labeling methods that directly use predicted results as pseudo labels not considering confidence in the target domain for unsupervised domain adaptation in classification tasks \cite{long2013transfer,zhang2017joint}. 
On the other hand, Pei {\textit{et al}.} utilized soft labeling methods that consider the probability of predicted classes as confidence in the classification tasks \cite{pei2018multi}.

Selective pseudo-labeling is a strategy that selects high confident pseudo labels in the target data in order to avoid using inaccurate labels.
Many such methods have been proposed \cite{choi2019pseudo,chen2019progressive,wang2019unifying,fu2021unsupervised}. Our method is also categorized to this approach.
Selective pseudo-labeling methods take confidence in target samples into account as soft labeling methods do but in different ways. Chen {\textit{et al}.}, Wang {\textit{et al}.}, Fu and Li {\textit{et al}.} proposed a selective pseudo-labeling method that utilizes the similarity distance from the class center via clustering \cite{chen2019progressive,wang2019unifying,fu2021unsupervised}. They selected samples in the target domain close to the class center. Similarly, Choi {\textit{et al}.} proposed a selective pseudo-labeling method that uses a curriculum learning based on the density of the cluster \cite{choi2019pseudo}. This method took density in clustering into consideration as confidence and selected samples with high-density step by step. However, all of these methods are for classification tasks, not detection tasks.
In classification tasks, features or confidence obtained by a classifier, which is directly related to the task, is directly used to select pseudo labels. In contrast, since the heatmap predictor could not produce confidence, we additionally introduce the Bayesian discriminator to identify correct heatmaps. In addition, we introduce curriculum learning to further improve label selection.

\subsubsection{Domain adaptation in object detection}
Several pseudo-labeling methods have been proposed for detection tasks \cite{kim2019self,d2020one,roychowdhury2019automatic,soviany2021curriculum}. RoyChowdhury {\textit{et al}.} utilized a self-training method that automatically computes pseudo labels from video data \cite{roychowdhury2019automatic}. Specifically, they used the temporal consistency between adjacent frames to track detection by the model trained with the source. Kim {\textit{et al}.} proposed a self-training one-stage object detector that utilizes pseudo labels to adapt to the target domain \cite{kim2019self}. Soviany {\textit{et al}.} proposed a curriculum self-paced learning approach for object detection \cite{soviany2021curriculum}. They used cycle-GAN to produce images that resemble the target images and introduced an image difficulty score (the number of detected objects divided by their average bounding box area). This method selected pseudo labels in order from easy to difficult by referring to an image difficulty score, which is a form of curriculum learning. 
All of these methods assume that the ground truth labels are given by bounding boxes, and they used the confidence produced by the classification network for selecting pseudo labels. 
However, cells are densely distributed, and they move with non-rigid deformation. Then, the bounding boxes of cells often overlap. In addition, the annotation for a bounding box label is more expensive than a point-level annotation.
In this paper, we propose a domain adaptation method for heatmaps that can represent the positions of cells from the point-level annotation.

\begin{figure}[t]
    \centering
    \includegraphics[width=\columnwidth]{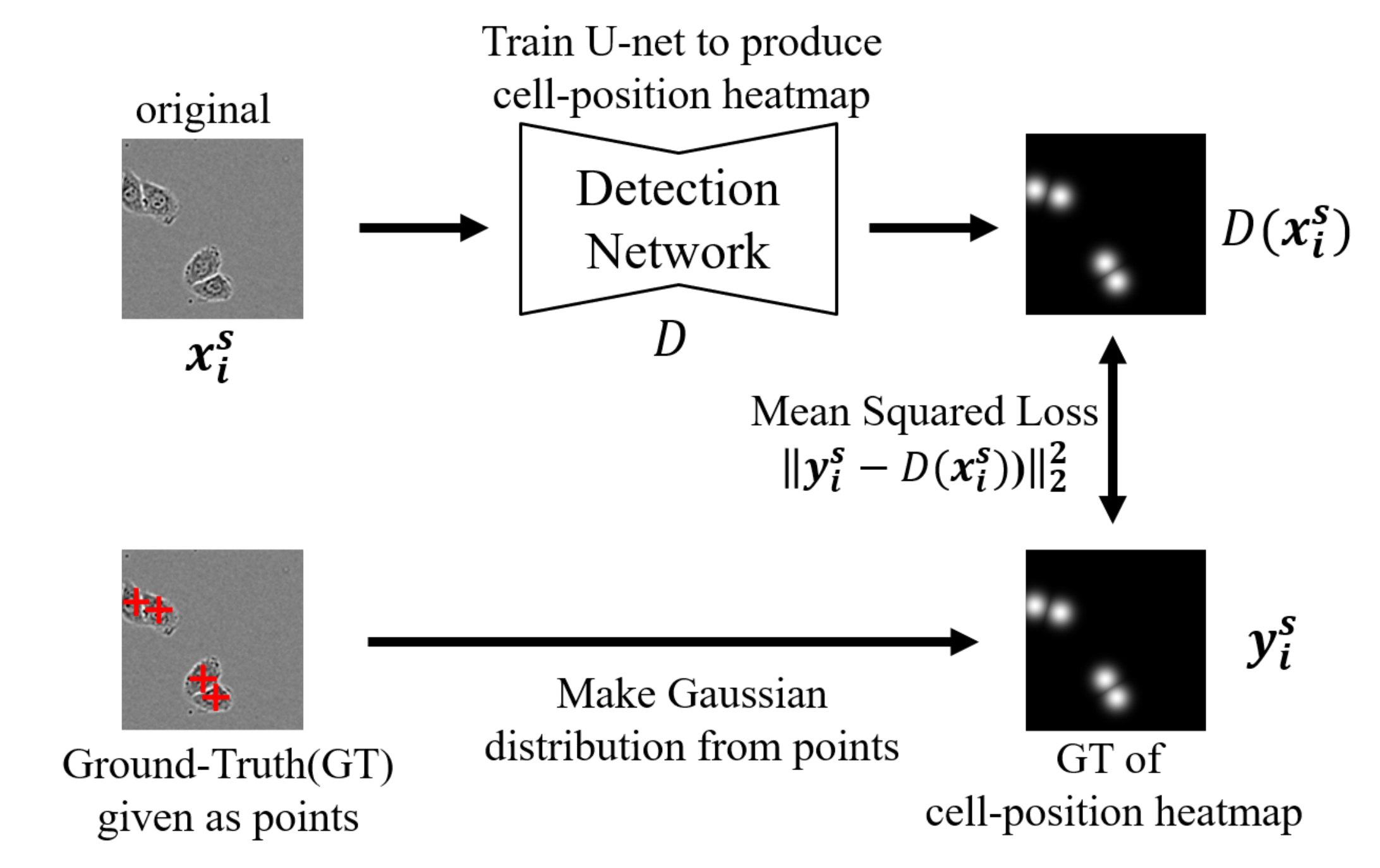}
    \caption{Training detection network. Initially, cell-position heatmap is generated from cell centroid position. Then, detection network is trained with MSE loss between $D(\bm{x}_i^s)$ and $\bm{y}_i^s$.}
    \label{fig:Detection network}
\end{figure} 

\begin{figure*}[t]
    \centering
    \includegraphics[width=\linewidth]{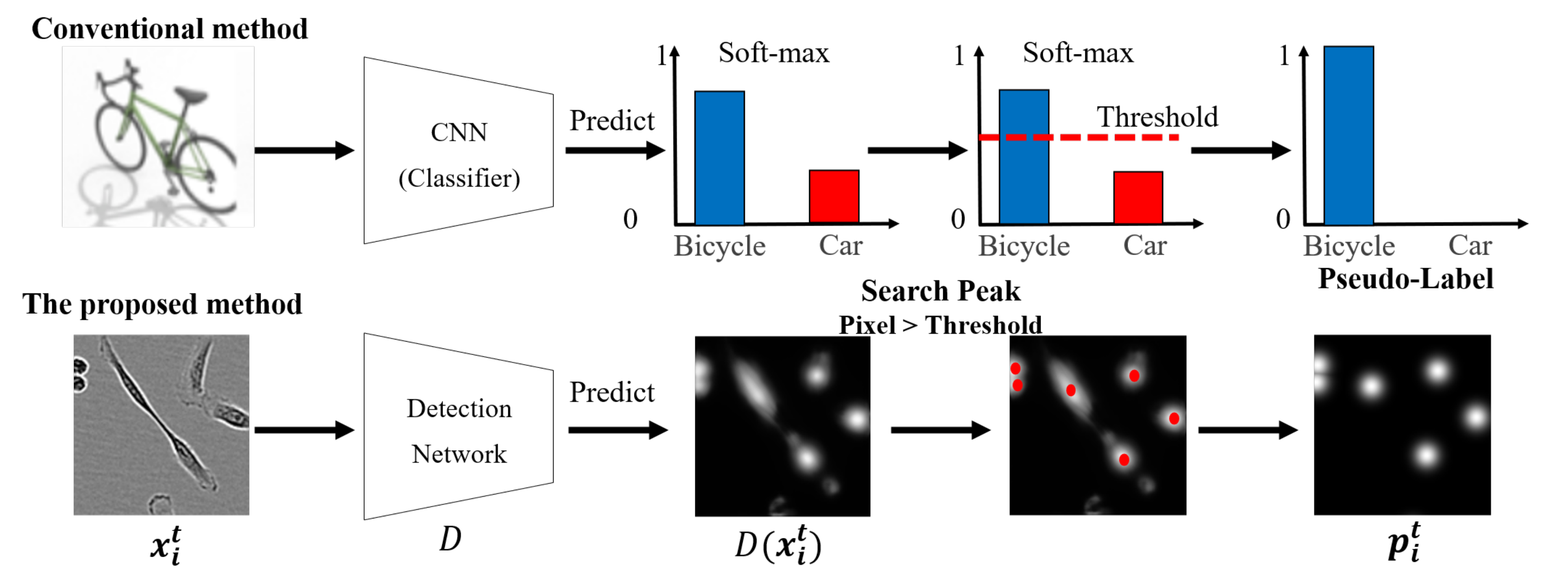}
    \caption{Generating a pseudo-cell-position heatmap. The top row illustrates the conventional pseudo-labeling method for classification tasks. The bottom row is the proposed method. First, we find peak positions using a threshold. Then, regenerate a clear Gaussian distribution from peak positions.}
    \label{fig:Pseudo labeling}
\end{figure*}

\section{Domain extension in cell detection}
In unsupervised domain adaptation, we are given a set of labeled source images $\mathcal{D}_s=\{(\bm{x}_i^s, \bm{y}_i^s)\}_{i=1}^{m_s}$ and a set of unlabeled target images $\mathcal{D}_t = \{\bm{x}_i^t\}_{i=1}^{m_t}$, where $\bm{x}_i^s$, $\bm{x}_i^t$ are the input images, ${m_s}$, ${m_t}$ are the number of data in the source and target domains, respectively, $\bm{y}_i^s$ is the corresponding ground truth of the heatmap, in which the peaks show the cell centroid positions (the details are described in Section \ref{sec:detection}).
Our aim is to train the model using the source and target data so that it achieves high performance in both domains.

As shown in Fig. \ref{fig:overview_of_method}, the overall network architecture is mainly composed of a detection network $D$, and a Bayesian discriminator $B$.
Initially, the entire images in the source domain and the target domain are separated into patch images by a simple sliding window operation to make it easy to handle cells with different shapes in an entire image.  We randomly selected the patch image for labeled data. The patch image may contain several cells. The proposed method consists of five consecutive stpdf. In step 1, using source data $\mathcal{D}^s$, we train the detection network $D$ that estimates the cell-position heatmap \cite{nishimura2019weakly} and the Bayesian discriminator $B$ that classifies whether generated pseudo-cell-position heatmap is correct or not. In step 2, $D$ estimates the prediction maps for $\bm{x}_i^t$ ($i=1,...,m_t$), in which the predicted cell-position heatmap $D(\bm{x}_i^t)$ may have a distorted shape (non-Gaussian) even if the peak position is correct, as shown in Fig. \ref{fig:overview_of_method}. In step 3, pseudo-cell-position heatmaps (pseudo ground truth) $\bm{p}_i^t$ with clear Gaussian shapes are generated on the basis of $D(\bm{x}_i^t)$. In step 4, given pairs of the original $\bm{x}_i^t$ and its $\bm{p}_i^t$ as inputs (these are concatenated as channels), $B$ predicts an uncertainty score if the inputted pseudo-cell-position heatmap $\bm{p}_i^t$ is correct for the original $\bm{x}_i^t$ and selects a pseudo-cell-position heatmap that it is confident of as a candidate training data. In step 5, the proposed method uses curriculum learning selecting from easy to hard pseudo-cell-position heatmaps in order. Finally, the selected easy pseudo-cell-position heatmaps are used as additional training data for $D$ and $B$. This process is iterated on the target domain data $\mathcal{D}^t$.

\subsection{Cell detection with cell-position heatmap (step 1 and 2)} \label{sec:detection}
Object detection tasks often use a bounding box as the ground truth for localizing objects. However, it is known that the bounding boxes do not work well for cell detection because cells have complex shapes, and a large number of them may appear in an image; individual bounding boxes may often contain two or more cells under such conditions. Instead, we use a cell-position heatmap, which was shown to have good cell-detection performance \cite{nishimura2019weakly}. Fig. \ref{fig:Detection network} shows the process of training the detection network. 
Given a set of annotated cell centroid positions for an image $\bm{x}_i^s$, the ground truth of the cell-position heatmap $\bm{y}_i^s$ is generated so that a cell centroid positions are at the peaks of Gaussian distributions in the map, as shown in Fig. \ref{fig:Detection network}. We generate each Gaussian distribution $\bm{y}_{i,n}^s$ as follow:
\begin{equation}
    \label{eq:indlikelihhod}
    \bm{y}_{i,n}^s(\vec{u}) = \exp \left(- \frac{||\vec{u} - \vec{z}_n ||^2_2}{\sigma^2} \right),
\end{equation} 
where $n$ is each cell, $\vec{u}$ indicates the position coordinate, $\vec{z}_n$ indicates the ground truth position of the $n$-th cell, and $\sigma$ adjusts the Gaussian blur as a hyper-parameter. Then, we generate the cell-position heatmap $\bm{y}_i^s=\max_n{\bm{y}_{i,n}^s}$ as the ground truth by maximizing $\bm{y}_{i,n}^s$. Finally, to train the detection network $D$, we use the mean of the squared error loss function (MSE) between the predicted map $D(\bm{x}_i^s)$ and the ground truth (cell-position heatmap) $\bm{y}_i^s$. Finally, the trained the detection network $D$ predicts the heatmap $D(\bm{x}_i^t)$ from the input $\bm{x}_i^t$ in the target domain.

\subsection{Pseudo-labeling for cell detection (step3)}
In general, semi-supervised classification tasks use pseudo-labeling to deal with a lack of training data. They use soft-max scores, which are the probability of the data belonging to a certain class, as a confidence score like the top of the Fig. \ref{fig:Pseudo labeling}. If the soft-max score is higher than the threshold probability for a class, that class is used as a pseudo label. 
This pseudo-labeling strategy can not be directly used for pseudo label selection from the estimated heatmap since each image contains multiple cells, where multiple Gaussian distributions appear in the heatmap.

To deal with this problem, we propose a pseudo-labeling method for estimating heatmaps in cell detection (multi-object detection) tasks.
Fig. \ref{fig:Pseudo labeling} (bottom) shows our pseudo-labeling method for cell detection.
In the target domain, there are cells with various appearances, and some of them are similar to the cells in the source domain. The estimated heatmaps for very similar cells show clear Gaussian distributions, while those for moderately similar cells may have a distorted Gaussian distribution, even though its peak positions are almost correct and those for non-similar cells have no response or a noisy distribution (the peak positions are not correct).
We consider that if the re-trained network can estimate a heatmap of moderately similar cells showing a clear Gaussian distribution, the number of moderately similar cells increases in the next iteration since such cells are moderately similar to the cells whose estimation was improved by the re-trained network.

To achieve this, we generate the pseudo-cell-position heatmap $\bm{p}_i^t$ on the basis of the prediction map $D(\bm{x}_i^t)$ so that detected positions in the predicted cell-position heatmap are at the peaks of clear Gaussian distributions in the same manner as in step 1. Here, if the peak of prediction map $D(\bm{x}_i^t)$ is higher than a threshold $th_d$, the peak position is detected (see the red points in Fig. \ref{fig:Pseudo labeling}). This idea is similar to the conventional pseudo-labeling in classification tasks. 

The generated pseudo-cell-position heatmaps are regarded as candidate pseudo labels.
As discussed above, some cells in the target domain are not similar to the cells in the source domain; thus, the pseudo heatmaps may contain incorrect positions.
In order to select the 'correct' pseudo labels for re-training, the following stpdf are performed.

\begin{figure}[t]
    \centering
    \includegraphics[width=\columnwidth]{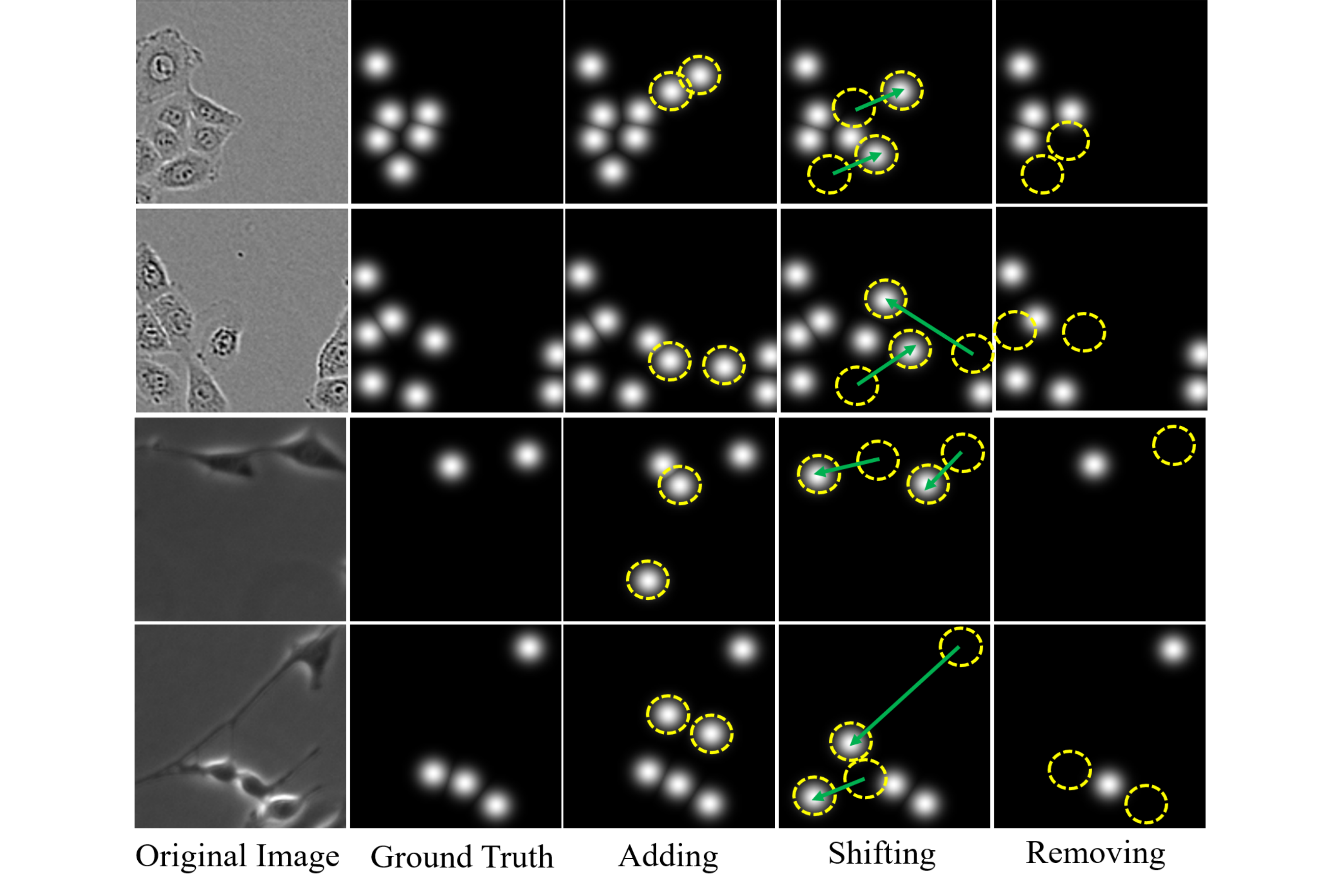}
    \caption{Examples of negative samples. Yellow circles indicate randomly added, shifted, and removed individual heatmaps. Green arrows indicate moving directions from the origin.}
    \label{fig:Negative samples}
\end{figure}

\subsection{Bayesian discriminator and curriculum learning (Stpdf 4 and 5)}
If the pseudo-cell-position heatmaps $\mathbf{P}^t=\{\bm{p}_i^t\}_{i=1}^{m_t}$ contain many incorrect labels, the performance of the detection network may be affected. To select confident pseudo-cell-position heatmaps, we introduce a Bayesian CNN $B$ and curriculum learning.

\subsubsection{Selection based on uncertainty}
Bayesian CNN $B$ estimates the uncertainty in whether the estimated detection result is correct or not.
To represent this model uncertainty, we leverage a dropout-based approximate uncertainty inference \cite{gal2015bayesian, gal2016dropout}, in which the model uncertainty can be estimated using the estimation of samples from the posterior distribution.

First, we explain the general formulation of the Bayesian CNN for classification \cite{gal2016dropout}.
Given the pair of the original data $X$ and its ground truth $Z$, the weights $W$ of the network is trained. The posterior distribution $p(W\mid X,Z)$ is approximated with the variational distribution $q(W)$ over the model weights by minimizing the Kullback-Leibler (KL) divergence as follow:
\newcommand{\argmin}{\mathop{\rm arg~min}\limits}
\begin{equation}
    q^{\ast}(W) = \argmin_{q(W)}KL(q(W)\parallel p(W\mid X,Z)).
\end{equation}
The model with dropout learns the distribution of weights while training. Then, we sample the posterior weights of the model by using the dropout at the test for the model to find the predictive distribution over output. \cite{gal2015bayesian, gal2016dropout}. 
The expectation of confidence can be computed by the average of the estimation results by the model with different dropouts as follows:
\begin{align}
\mathbb{E}_{q} = \frac{1}{T}\sum_{t=1}^{T}p(\,\widehat{z}^{\,\ast}\mid\widehat{x}^{\,\ast},\widehat{w}_{t})),
\end{align}
where $T$ is the number of sampling (the number of networks with dropout), $\widehat{x}^{\,\ast}$ and $\widehat{z}^{\,\ast}$ are the input and output from the model and $\widehat{w}_{t}$ are weights sampled from the distribution over the model's weights. 
Model uncertainty in a classification task can be quantified by the entropy of the expectation \cite{gal2016dropout,corbiere2021confidence}.

We use this Bayesian CNN to select the correct pseudo heatmaps from the candidates obtained in the step 3.
The input of the network in the inference is a set of the original image and the corresponding pseudo-cell-position heatmap $\{(\bm{x}_i^t, \bm{p}_i^t)\}_{i=1}^{m_t}$, and the output is a label $Z=\{0,1\}$ with uncertainty. If $\bm{p}_i^t$ seems to be a correct label for the original image $\bm{x}_i^t$, $Z$ takes 1; otherwise, 0. If the predicted label is confident, it produces lower uncertainty; otherwise, higher uncertainty. To train $B$ in the initial iteration, we deliberately generate negative samples (the incorrect ground truth) from the ground truth of $\bm{x}_i^s$ (Fig. \ref{fig:overview_of_method}) by randomly adding, removing, or shifting the Gaussians in the ground truth of $\bm{x}_i^s$. 
This negative sample generation is performed in each iteration on the basis of the selected pseudo labels. Fig. \ref{fig:Negative samples} shows negative samples for each pattern. Here, to avoid overlap of the Gaussian distributions by augmentation, we shifted or added Gaussians to the regions that are distant (15 pixels) from the ground truth of the existing cell centroids. Then, the model is trained with the Cross-Entropy loss to classify whether pseudo-cell-position is correct or not. 
In inference, we apply the network to a single input $T$ times with different random dropouts and on the basis of these $T$ results obtain the predicted labels with uncertainties for all image pairs of $\{(\bm{x}_i^t, \bm{p}_i^t)\}_{i=1}^{m_t}$. 
We then select the samples of the top $th_u$ of higher certainty as pseudo-cell-position heatmaps.

\begin{figure}[t]
    \centering
    \includegraphics[width=0.7\columnwidth]{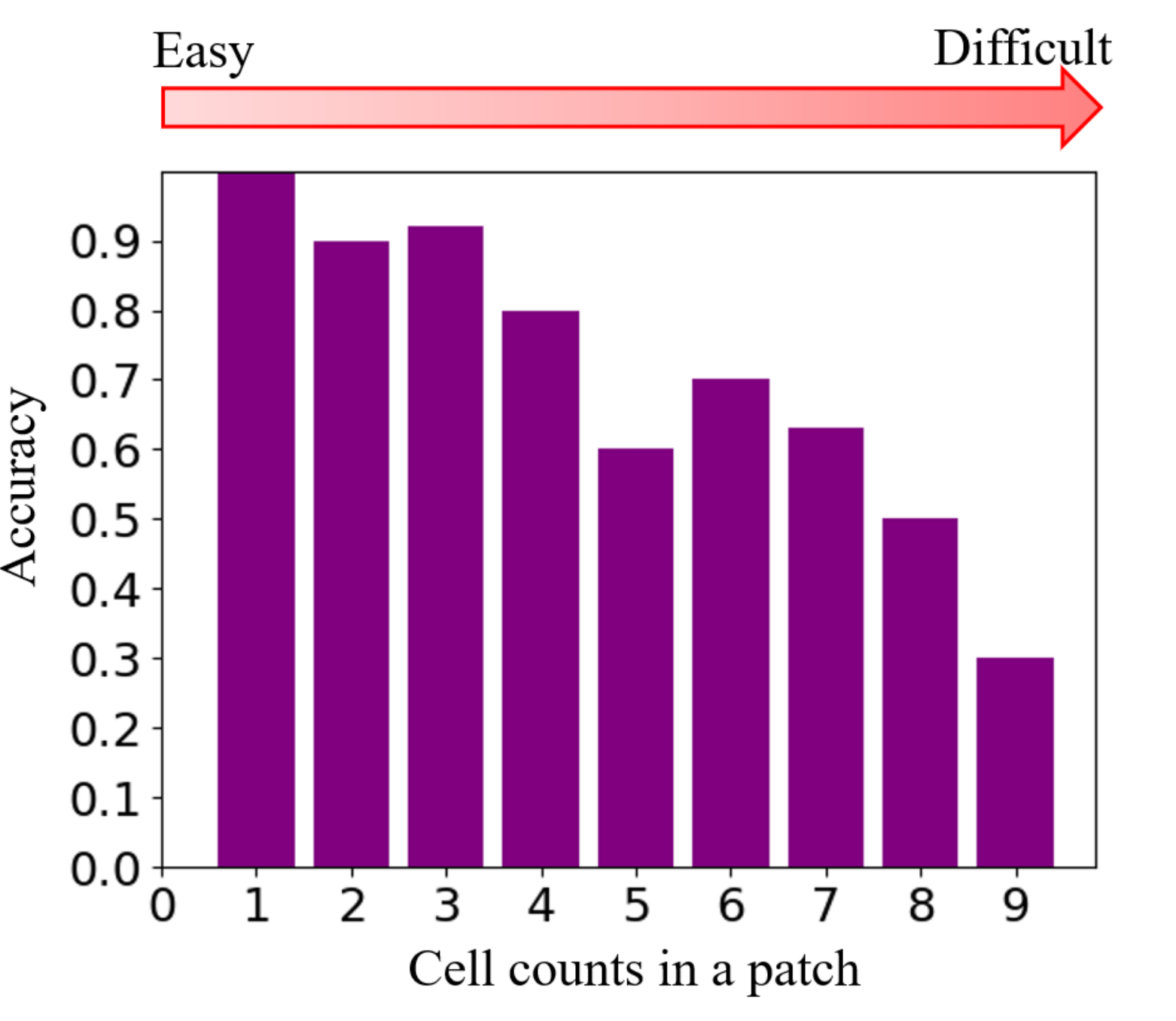}
    \caption{Relationship between correctness (vertical axis) and the number of cells (horizontal axis) in the heatmap selected by the Bayesian discriminator. There was tendency that accuracy decreases as the number of cells in a patch increases.}
    \label{fig:Curriculum_learning_ablation_study}
\end{figure} 

\subsubsection{Selection based on curriculum learning} 
Even if the Bayesian discriminator selects confident pseudo labels, they still contain some noisy labels.
This leads to an adverse effect on the performance of the detection network and the Bayesian discriminator.
We consider that these incorrect pseudo labels cause especially adverse effects at the beginning of learning.

As a further improvement from our preliminary work \cite{cho2021cell}, we introduce curriculum learning that first selects easier cases and then gradually covers difficult cases step-by-step for selecting correct pseudo labels at the beginning of learning.
In a cell detection task in the target domain, cell density (the number of cells in a local patch image) can be considered as the difficulty of detection. When the cell density is sparse ({\it i.e.}, the number of cells in a patch image is fewer), the detection is easier.
Fig. \ref{fig:Curriculum_learning_ablation_study} shows the performance of the Bayesian discriminator on selected pseudo-cell-position heatmaps in the first iteration, where the horizontal axis is the number of detected cells in the detection result of a patch image. Indeed, patches with fewer cells have more correct pseudo-cell-position heatmaps.

\begin{figure}[t]
    \centering
    \includegraphics[width=0.83\columnwidth]{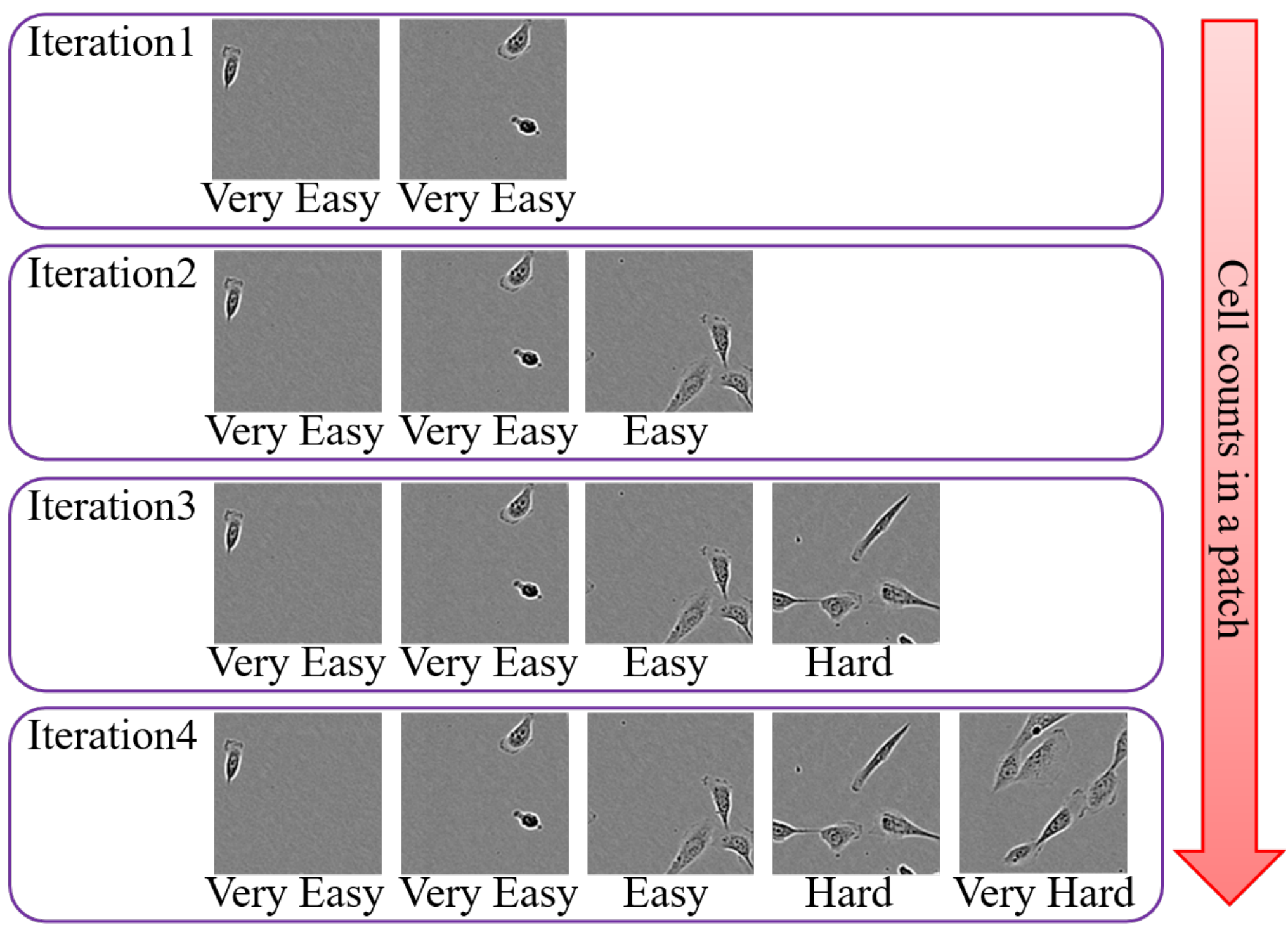}
    \caption{Curriculum learning based on the number of cells. The number of cells in a patch gradually increases with iteration in learning.
    }
    \label{fig:Curriculum_learning2}
\end{figure}

On the basis of this observation, we design a curriculum learning based on the number of cells in a patch, as shown in Fig. \ref{fig:Curriculum_learning2}. In the first iteration, we select the patch images that contain only one or two cells from the candidate pseudo labels selected by Bayesian discriminator as pseudo labels and then increment the number of cells by $N_c$ in each subsequent iteration.
Here, we use a number based on the estimated results since we cannot get the actual number of cells in a patch.

The selected pseudo-cell-position heatmaps are used as the training data in the target domain for $D$ and $B$. Even if the signals of a cell in the initially estimated cell-position heatmap have a non-Gaussian shape, cells that have a similar appearance to the pseudo labels can be detected with a clear Gaussian shape by the re-trained $D$ using the pseudo-cell-position heatmap. $D$ will be incrementally improved as this pseudo-labeling process iterates.

\section{Experiment}
To evaluate the effectiveness of our method on different domains, we experimented with 14 combinations of domains using 2 datasets, where one dataset contains 4 conditions (12 pairs) and the other has 2 conditions (2 pairs).
The experiments focused on the domain shift from differences in cell appearance due to different culture conditions since our method assumes that the source and target data have some intersection, as shown in Fig. \ref{fig:our idea}. Therefore, we selected two datasets that contain images under several culture conditions.\footnote{We here note that current domain adaptation methods for general objects \cite{french2017self,liang2020we} also have a similar limitation. Therefore, their evaluation does not contain some pairs of domains that have large differences. Ex) In hand-writing image datasets, domain adaptation from MNIST \cite{lecun1998gradient} or USPS \cite{hull1994database} to SVHN \cite{netzer2011reading} was not performed in any state-of-the-art papers due to the limitation \cite{french2017self,liang2020we}.}

\begin{figure*}[t]
    \centering
    \includegraphics[width=\textwidth]{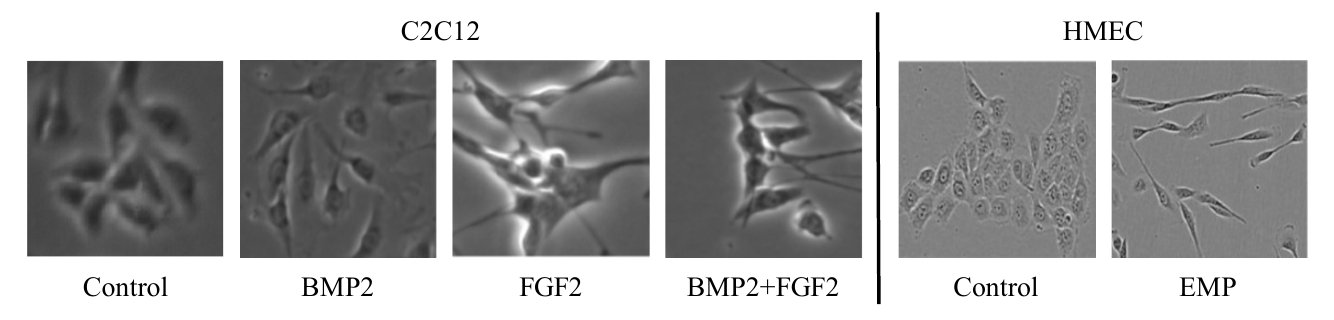}
    \caption{Cell images on all conditions in all datasets. The left side is the C2C12 dataset, and the right side is the HMEC datset. The appearance of cells are significantly different among the culture conditions.}
    \label{fig:Datasets}
\end{figure*}

\begin{table*}[t]
    \caption{Detection performance (F-score) on different domain. 'S' and 'T' indicate the source and target domain, respectively. 'B' (BMP2), 'C' (Control), 'E' (EMP2),'F' (FGF2) and 'B+F' (BMP2+FGF2) indicate the culture conditions. We used F1-score to evaluate the detection performance.}
    \label{tab1}
    \resizebox{\textwidth}{!}{
    \begin{tabular}{cl|cc|ccccccc}
    \multicolumn{1}{l}{Data} & \multicolumn{1}{l|}{S to T} & \multicolumn{1}{l}{S→S} & \multicolumn{1}{l|}{Ours on S} & \multicolumn{1}{l}{S→T} & \multicolumn{1}{l}{\citeauthor{vicar2019cell}} & \multicolumn{1}{l}{\citeauthor{moskvyak2021semi}} & \multicolumn{1}{l}{\citeauthor{haq2020adversarial} w/o AE} & \multicolumn{1}{l}{\citeauthor{haq2020adversarial}} & \multicolumn{1}{l}{Ours w/o curriculum} &\multicolumn{1}{l}{Ours}\\ \hline\hline
    \multicolumn{1}{c}{\multirow{12}{*}{C2C12}} & C→F& 0.850 & \textbf{0.867} & 0.705& 0.612& 0.642& 0.755& 0.767& 0.807 & \textbf{0.832}\\
    \multicolumn{1}{c}{} & F→C & 0.800 & \textbf{0.833} & 0.684& 0.700& 0.642& 0.766& 0.771& 0.845& \textbf{0.855} \\                    
    \multicolumn{1}{c}{}& C→B & 0.850 & \textbf{0.862} & 0.709& 0.584& 0.761& 0.779& 0.795& 0.820 & \textbf{0.859} \\
    \multicolumn{1}{c}{} & B→C & 0.885 & \textbf{0.941} & 0.756& 0.700& 0.775& 0.843& 0.848& 0.860 & \textbf{0.869} \\
    \multicolumn{1}{c}{} & C→B+F & 0.850& \textbf{0.879} & 0.780& 0.738& 0.770& 0.773& 0.787& 0.836 & \textbf{0.850}\\
    \multicolumn{1}{c}{}& B+F→C & \textbf{0.897} & 0.890 & 0.815& 0.700& 0.679& 0.800& 0.808& \textbf{0.851} & 0.847\\
    \multicolumn{1}{c}{}& F→B  & 0.800 & \textbf{0.835} & 0.742& 0.584& 0.721& 0.761& 0.757& \textbf{0.817} & 0.795\\
    \multicolumn{1}{c}{}& B→F   & 0.885 & \textbf{0.899} & 0.632& 0.612& 0.648& 0.766& 0.771& 0.798 & \textbf{0.850}\\
    \multicolumn{1}{c}{}& F→B+F  & 0.800 & \textbf{0.833} & 0.850& 0.738& 0.828& 0.832& 0.844& \textbf{0.873} & 0.872\\
    \multicolumn{1}{c}{}& B+F→F & 0.897 & \textbf{0.905} & 0.830& 0.612& 0.764& 0.788& 0.785& \textbf{0.849} & 0.833\\
    \multicolumn{1}{c}{}& B→B+F& 0.885 & \textbf{0.919} & 0.792& 0.738& 0.839& 0.863& 0.867& 0.890 & \textbf{0.913}\\
    \multicolumn{1}{c}{}& B+F→B  & 0.897 & \textbf{0.900} & 0.780& 0.584& 0.572& 0.789& 0.825& \textbf{0.894} & 0.889\\ \hline
    \multicolumn{1}{l}{\multirow{2}{*}{HMEC}}& C→E & 0.941 & \textbf{0.962} & 0.768& 0.797& 0.798& 0.851& 0.849& 0.875 & \textbf{0.904} \\
    \multicolumn{1}{l}{}& E→C& 0.941 & \textbf{0.971} & 0.939& 0.857& 0.835& 0.924& 0.921& \textbf{0.959} & 0.956 \\\hline
    \multicolumn{2}{c|}{Average}& 0.870& \textbf{0.893} & 0.770 & 0.683& 0.734& 0.806 & 0.814& 0.855& \textbf{0.866}              
    \end{tabular}%
    }
\end{table*}

\subsection{Datasets}
\noindent
\textbf{C2C12:} This is open-source dataset \cite{eom2018phase} consisting of myoblast cells captured by phase-contrast microscopy at a resolution of $1040 \times 1392$ pixels and cultured under four different conditions; 1) Control (no growth factor), 2) FGF2 (fibroblast growth factor), 3) BMP2 (bone morphogenetic protein), 4) BMP2+FGF2 (fibroblast growth factor + bone morphogenetic protein). As shown in Fig. \ref{fig:Datasets}, the cell appearances differ between conditions. For example, the cells in FGF2 are extremely elongated or shrunken, and they severely touch, while the ones in BMP2 are much larger than those in Control.
We separated patches of $128 \times 128$ pixels from the full images of the datasets. The ground truth was given to only 24 patches (approximately a quarter of one entire image) in each condition. In the evaluation of domain adaptation experiments, we used one sequence as unlabeled data (100 entire images) and the other sequence as test data (100 entire images) under different conditions from the training data. In the evaluation of the same domain experiments, we used one sequence as unlabeled data (100 entire images) and the other sequence as test data (100 entire images) under the same conditions from the training data. Note that we did not use validation samples to tune hyper-parameters because of the protocol of unsupervised domain adaptation.

\noindent
\textbf{HMEC:} This dataset contains images of human mammary epithelial cells (HMEC) captured by phase-contrast microscopy at a resolution of $1272 \times 952$ and cultured under two conditions; 1) Control (no stimulus) and 2) EMT (epithelial-mesenchymal transition) \cite{nieto2016emt}. Example images for each condition are shown in Fig. \ref{fig:Datasets}, where the appearance of the cells in Control are round and those in EMP2 are elongated.
In the evaluation of all experiments, we used the same setting with C2C12. 

\begin{figure*}[t]
    \centering
    \includegraphics[width=\linewidth, left]{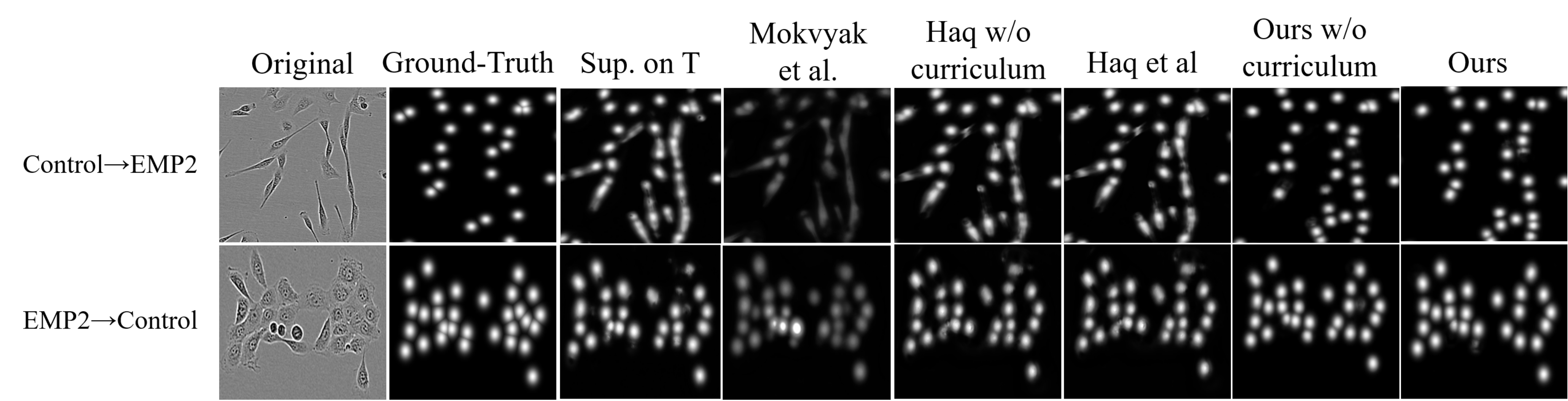}
    \caption{Examples of detection results on HMEC on different domains.}
    \label{fig:HMEC detection result}
\end{figure*}

\begin{table*}[t]
    \caption{Detection performance on the same domain.'Cond.' indicates the culture condition on each datset.}
    \label{tab2}
    \centering
    \begin{tabular}{cc|ccccc}
    \multicolumn{1}{l}{Data} & \multicolumn{1}{l|}{Cond.} & 
    \multicolumn{1}{l}{Sup.}&
    \multicolumn{1}{l}{\citeauthor{vicar2019cell}} & \multicolumn{1}{l}{\citeauthor{moskvyak2021semi}} &  \multicolumn{1}{l}{Ours w/o curriculum} &\multicolumn{1}{l}{Ours}\\ \hline\hline
    \multicolumn{1}{c}{\multirow{4}{*}{C2C12}} & C& 0.850 & 0.700 & 0.851& \textbf{0.888}& 0.882\\
    \multicolumn{1}{c}{} & B & 0.885 & 0.584 & 0.899& \textbf{0.924}& 0.920 \\          
    \multicolumn{1}{c}{}& F & 0.800 & 0.612 & \textbf{0.859}& 0.842& 0.838\\
    \multicolumn{1}{c}{}& B+F & 0.897 & 0.738 & 0.904& \textbf{0.922}& 0.918\\ \hline
    \multicolumn{1}{l}{\multirow{2}{*}{HMEC}}& C & 0.941 & 0.857 & 0.963& 0.961& \textbf{0.964} \\
    \multicolumn{1}{l}{}& E & 0.941 & 0.797 & 0.958& 0.961& \textbf{0.966} \\\hline
    \multicolumn{2}{c|}{Ave.} & 0.886 & 0.858 & 0.906 & \textbf{0.916} & 0.915          
    \end{tabular}%
\end{table*}

\subsection{Evaluation metrics}
\label{sec:Discriminator performance}
\textbf{Detection Performance:} To evaluate detection performance, we used F1-score (\cite{qu2019improving}) as the performance metric. It was measured by making a one-by-one matching between the detection results and the ground truth. Each detected cell point was assigned to one ground truth by minimizing the sum of the distances between the ground truth and the detection results using linear programming (\cite{qu2019improving}). If the distance between a detection result and the assigned ground truth was less than a threshold (10 pixels), we counted it as a True Positive (the threshold was determined based on the cell size.). Non-assigned detection results were counted as False Positive and non-assigned ground truth results were counted as False Negative.

\textbf{Discriminator Performance:}To evaluate how well the discriminator selected the correct pseudo labels, we computed image-level accuracy from the F1-score. We compared the pseudo label with the ground truth and defined pseudo labels that had no False Positive and False Negative as measured by F1-score to be correct. Pseudo labels that had a False Positive or a False Negative were defined as incorrect.

\subsection{Implementation details}
We implemented our method with Pytorch and the Bayesian CNN implementation in \cite{Javier2020}. We used the U-Net structure \cite{ronneberger2015u} as the detection network and the Resnet18 structure \cite{szegedy2017inception} as the Bayesian discriminator. We set $th_d$, $th_u$, $T$, $N_c$ and dropout rate to 100, 0.1, 10, 1 and 0.3 and iterated our method 5 times. Moreover, we used $\sigma=6$ to generate the ground truth and pseudo-cell-position heatmap. We trained the detection network and the Bayesian discriminator with the Adam optimizer and set the learning rate to 1.0 $\times$ $10^{-3}$ for 200 epochs each time. All datasets, including the original image and the ground truth, were normalized to values between 0 and 255. The same parameters were used in all of the experiments. 

\subsection{Compared methods}
The baseline (Sup.) of the proposed method is a supervised method \cite{nishimura2019weakly} that uses only training data from the source domain. We also compared the proposed method with four different methods; the unsupervised image-processing-based method proposed by Viscar {\textit{et al}.} that combined \cite{yin2012understanding} and distance transform \cite{vicar2019cell}, the semi-supervised learning method proposed by Moskvyak {\textit{et al}.} that used consistency of prediction and is not for domain adaptation \cite{moskvyak2021semi}, the Haq w/o AE method Haq {\textit{et al}.} that simply introduces adversarial domain adaptation method for cell segmentation \cite{haq2020adversarial}, and the domain adaptation method proposed by Haq {\textit{et al}.} that used an auto-encoder and adversarial learning for cell segmentation \cite{haq2020adversarial}. Note that we used the U-Net structure \cite{ronneberger2015u} as the detection network for these other methods. Additionally, we experimented with our method with curriculum learning (Ours) and without curriculum learning (Ours(w/o curriculum)) to check the effect of curriculum learning. Note that the training of these two methods used the same number of pseudo labels as the training data.

\begin{figure*}[p]
    \centering
    \includegraphics[width=0.9\linewidth]{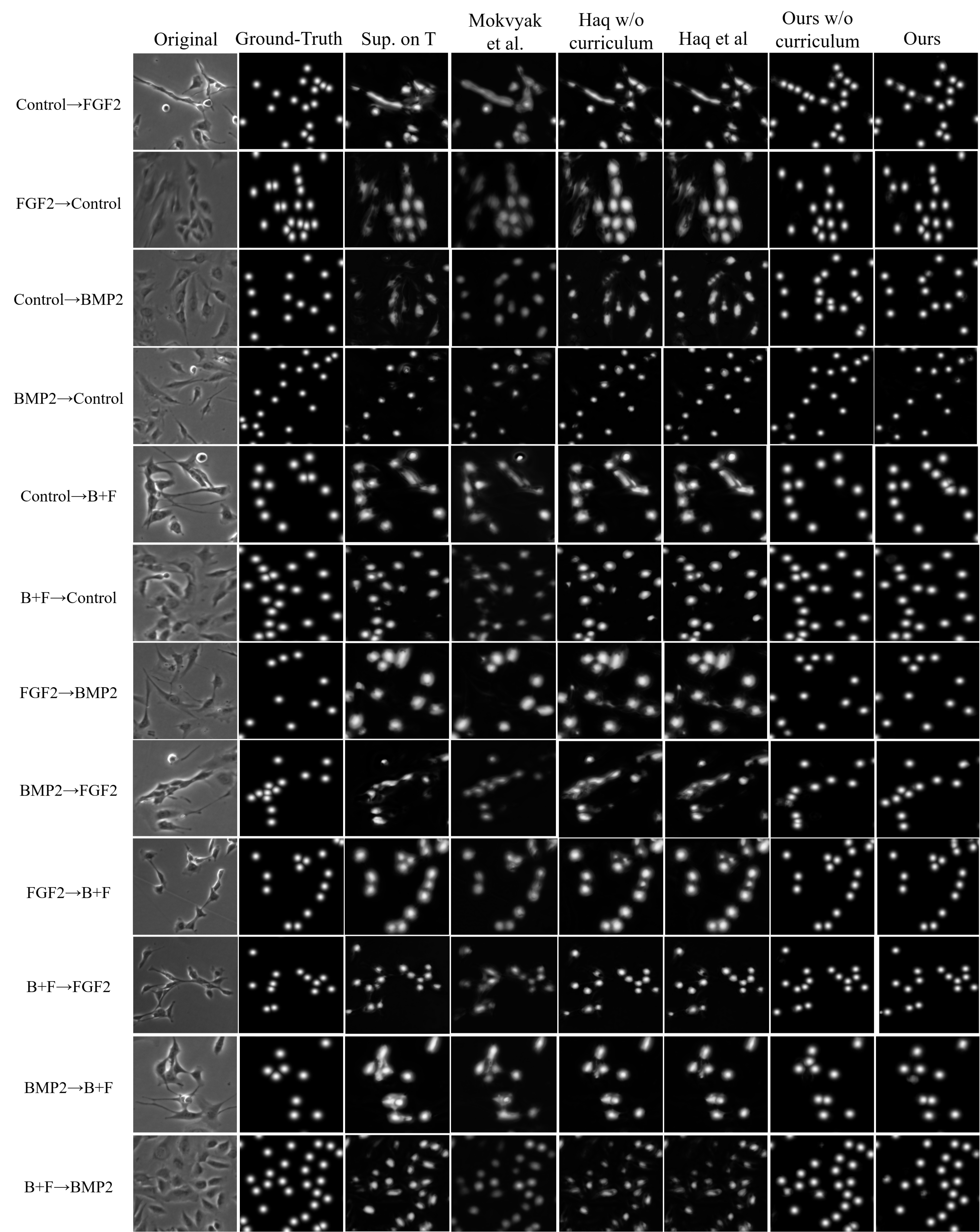}
    \caption{Examples of detection results on C2C12 where domains are different.}
    \label{fig:C2C12 detection result}
\end{figure*} 

\begin{figure}[t]
    \centering
    \includegraphics[width=\linewidth]{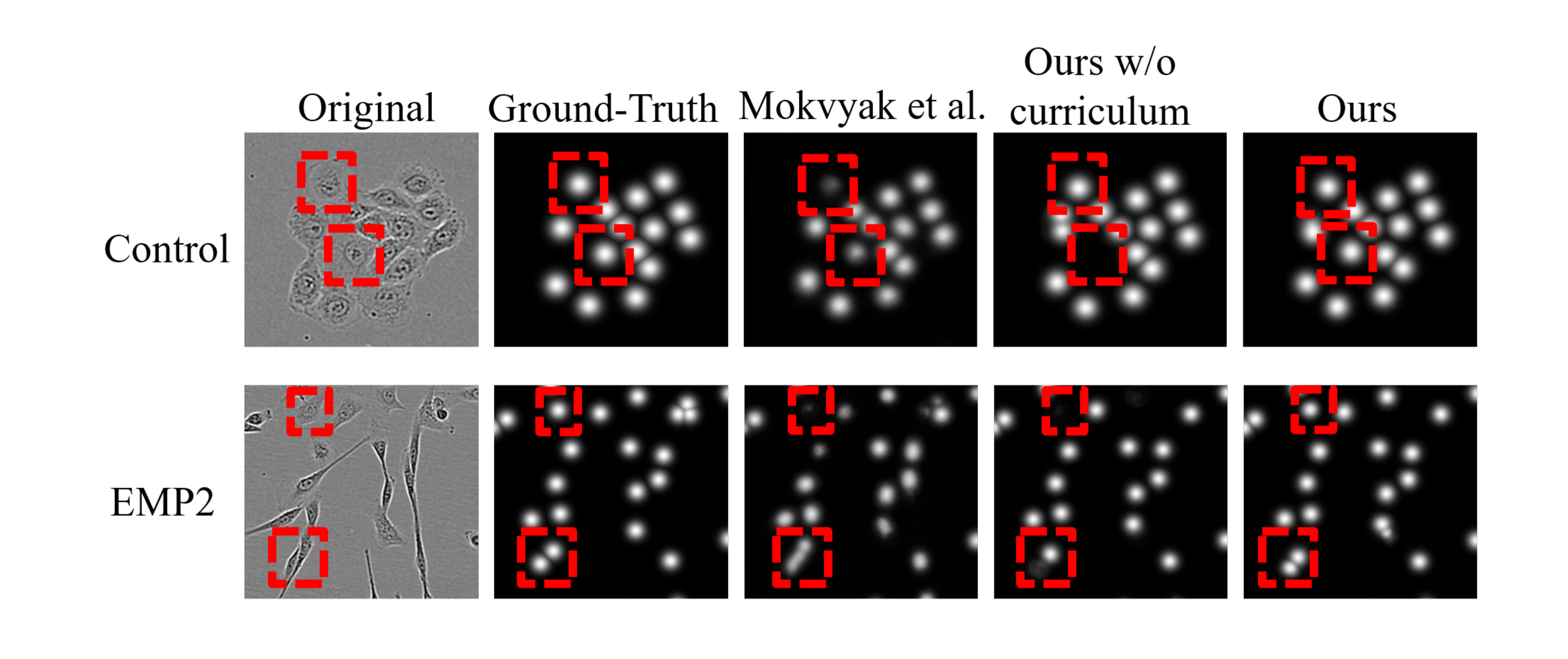}
    \caption{Examples of detection results on HMEC where domains are same.}
    \label{fig:HMEC detection result on same domain}
\end{figure} 

\begin{figure}[t]
    \centering
    \includegraphics[width=\linewidth]{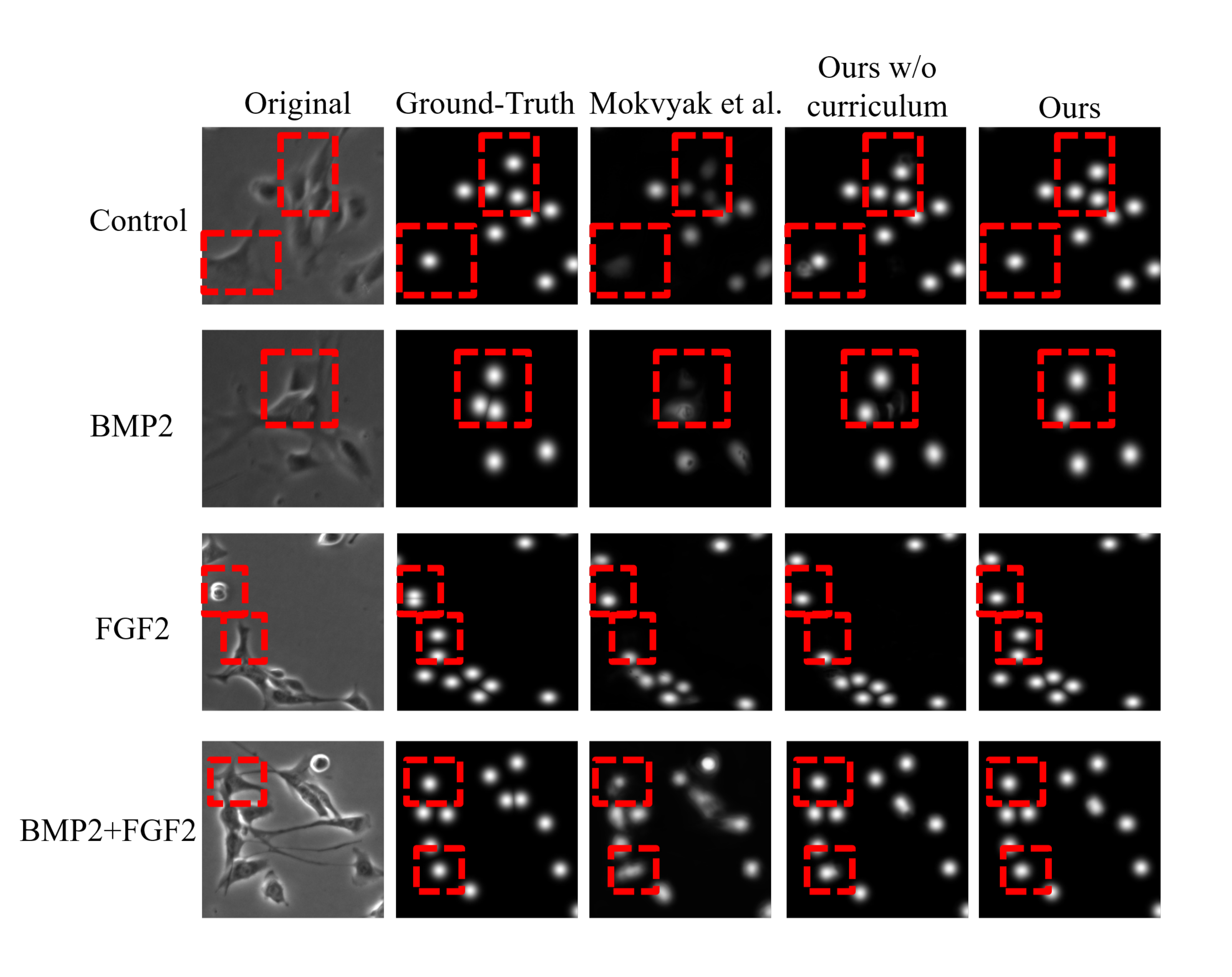}
    \caption{Examples of detection results on C2C12 where domains are the same.}
    \label{fig:C2C12 detection result on same domain}
\end{figure} 

\begin{figure}[t]
    \centering
    \includegraphics[width=0.9\linewidth]{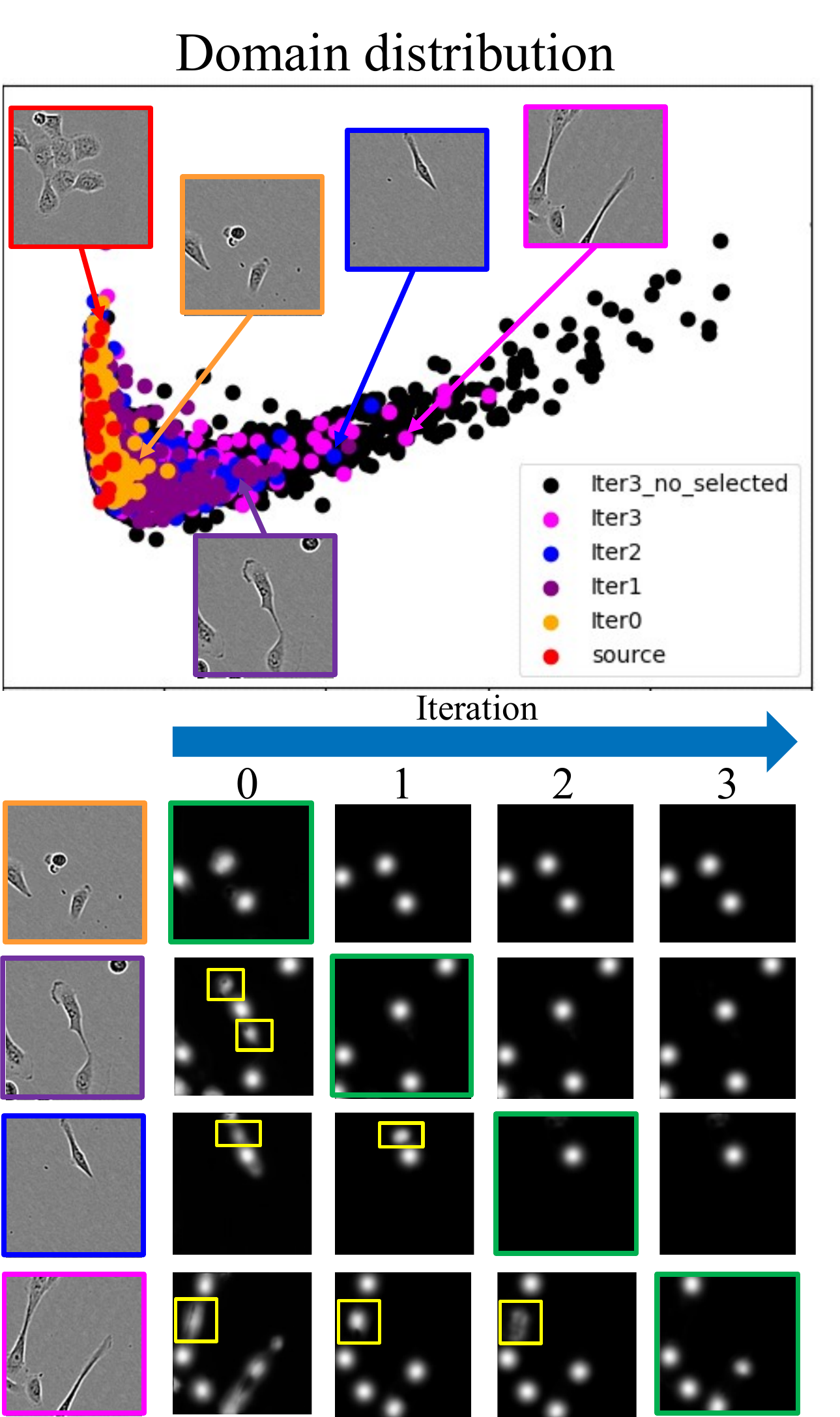}
    \caption{Visualization of the domain distribution in feature space and detection results at each iteration. Each color indicates the selected pseudo-cell-position heatmaps at each iteration. In the right figure, the yellow boxes indicate results that are mistakenly detected by the model. The green boxes represent the moment when pseudo-cell-position heatmaps were selected.}
    \label{fig:Domain distribution}
\end{figure} 

\subsection{Evaluation}
\textbf{Different domains:} Table \ref{tab1} shows the detection performances (F1-score) of the proposed method in the target domain and source domain for different domain settings. The source and target are denoted as S and T, and each culture condition is denoted by the first letter of the condition name. 'S→S' indicates the detection performance of the baseline when we evaluated this method on the other sequences in the same domain.
Note that the 'C→F' and 'C→B' in the 'S→S' column indicate both the training and test data are in the same domain 'C', and thus, their performances are the same. ‘S→T’ indicates the detection performance in the target domain when the model was trained with only the source domain. The detection performance of the baseline (S→T) was significantly lower than that of (S→S) in the same domain (average: -0.1). Although the Haq w/o AE and Haq's methods improved performance in the target domain more than the other methods (S→S, \cite{vicar2019cell} and \cite{moskvyak2021semi}) did, our methods (Ours w/o curriculum and Ours) outperformed all of these methods for all combinations. 

The (S→S) and (Ours. on S) columns in Table \ref{tab1} shows the detection performances on the additional test data in the source domain after our method was applied to the target domain. On almost all data, our method showed improved performance not only in the target domain but also in the source domain (average: +0.023). We consider that the pseudo-cell-position heatmaps for cells that have similar appearances to those in the source domain improved the detection performance of the proposed method in the source and target domain. Fig. \ref{fig:HMEC detection result} and Fig. \ref{fig:C2C12 detection result} show detection results for the test data in the target domains. As shown in Fig. \ref{fig:HMEC detection result} and Fig. \ref{fig:C2C12 detection result} baseline(S→T), the supervised method did not effectively estimate cell-position heatmaps, but the peaks in some were correctly detected. In contrast to this, our method could predict these maps with a clear Gaussian shape compared to other methods. These qualitative results support the idea of the proposed method.

\begin{figure}[t]
    \centering
    \includegraphics[width=\linewidth]{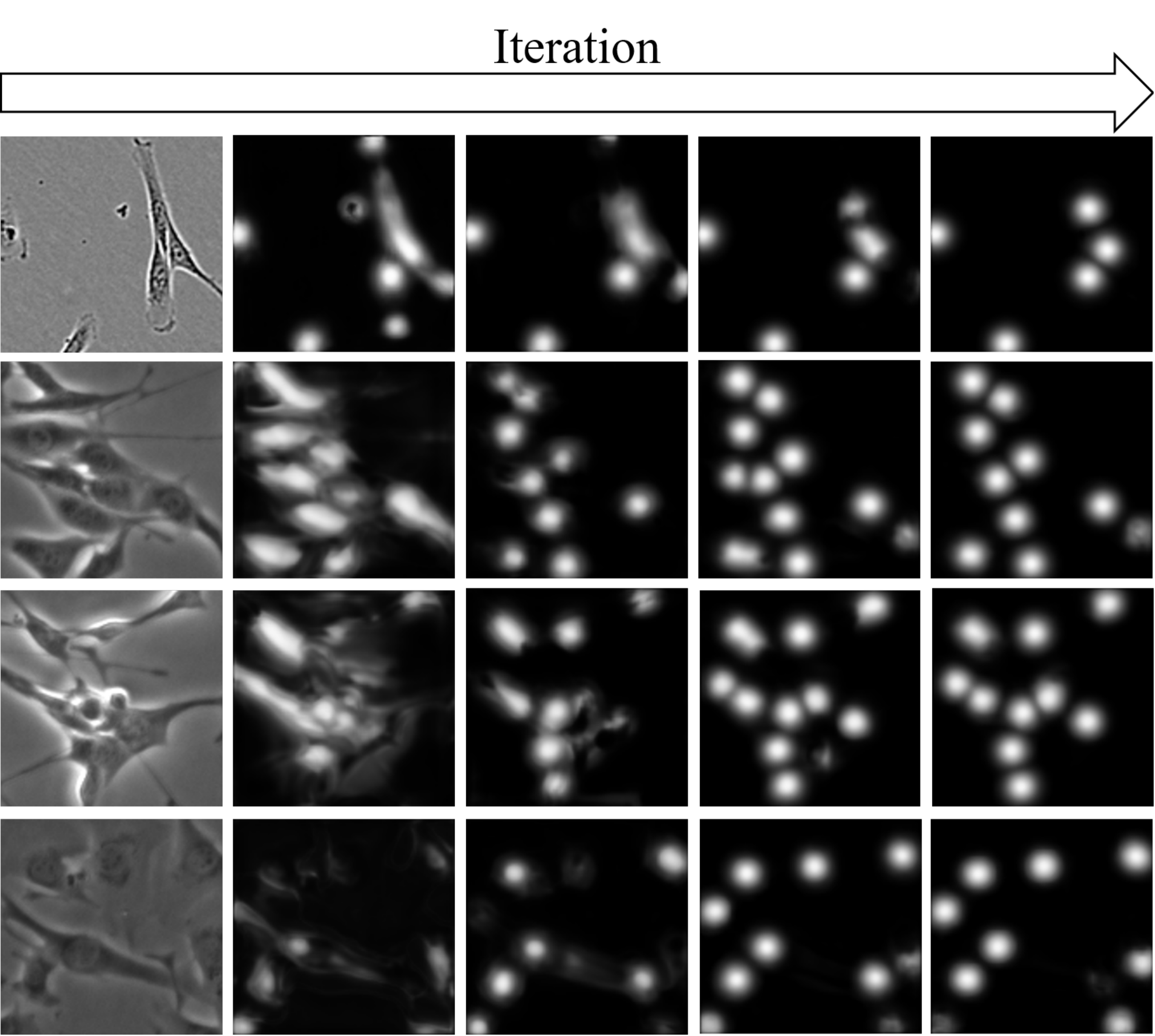}
    \caption{Examples of prediction maps on cells having complex shapes with iteration.}
    \label{fig:Prediction map changing with iteration}
\end{figure} 

\textbf{Same domains:} To evaluate the effectiveness of the proposed method as a semi-supervised learning, we experimented with our method and other methods in the same domain setting to confirm that our method can improve the detection performance from the few labeled and many unlabeled data. In this experiment, we used 24 patch images as labeled data and 8000 patch images (100 entire images) as unlabeled data for each condition. We conducted the evaluation in which all images were in the same domain. We did not evaluate Haq w/o AE and Haq’s methods since they are not for semi-supervised learning on the same domain. Indeed, in our experiment on the same domain (C→C), the performance of Haq’s method (F1-score: 0.843) fell below the baseline performance (F1-score: 0.850). Table \ref{tab2} quantitatively compares the detection performance in the same domain. Our method performed better for all domains compared with the baseline(Sup.), and it was better on average than Moskvyak's method \cite{moskvyak2021semi}, which was designed for semi-supervised learning. In the semi-supervised problem setup, curriculum learning could not further improve the detection performance, where the performance is comparative with w/o curriculum. We consider that the pseudo labels selected by the Bayesian discriminator from the same domain were more correct than those in the different domain because the features between labeled and unlabeled data in the same domain have similar distributions compared to those in different domains. Thus, the performance did not depend on the number of cells in the patch image. Fig. \ref{fig:HMEC detection result on same domain} and Fig. \ref{fig:C2C12 detection result on same domain} show detection results for the test data in the same domain. As shown in Fig. \ref{fig:HMEC detection result on same domain} and Fig. \ref{fig:C2C12 detection result on same domain} (red boxes), our method  detected cells amidst a high density of cells better than the other methods. These qualitative results show the effectiveness of the proposed method in the same domain.

\subsection{Observation}
\textbf{Domain distribution:} To examine the feature distribution of the source and target domain, we visualized the features extracted by the Bayesian discriminator by using T-SNE \cite{van2008visualizing}. The left image in Fig. \ref{fig:Domain distribution} shows the feature distribution of the source (control of HMEC) and the target domain (EMT) in the same feature space, where samples selected by the Bayesian discriminator are in a different color. In this distribution map, the cells in the source domain are round (red) and are distributed on the left side, whereas the cells in the target domain have various appearances (orange, purple, blue, magenta, and black) and are distributed from left to right. The colored box images are samples from this distribution. We can see that the Bayesian discriminator of the proposed method incrementally increased the pseudo patches in the feature distribution map as the iterations grow from left to right. The right image in Fig. \ref{fig:Domain distribution} shows examples of the predicted cell-position heatmaps in each iteration. In iteration 0, our method mistakenly detects the elongated cell located on the right side of the feature map as several cells. The cells far from the source domain in the distribution map were not selected in the early iterations, but the detection results of the cell-position heatmap improved in the later iterations and they were eventually selected when the prediction succeeded (Green box). These results indicate that the discriminator performed well in selecting the correct pseudo-cell-position heatmap and the proposed method incrementally extended the domains from the source to the target.

Fig. \ref{fig:Prediction map changing with iteration} shows the examples of prediction maps of the proposed method on cells having complex shapes with iteration. In the first iteration, estimated cell-position heatmaps have a similar appearance to the shapes of cells, and some of the peak positions on the prediction map are correct. In the next step, the prediction map has clearer Gaussian distributions by using a pseudo-cell-position heatmap. As shown in the last row in Fig. \ref{fig:Prediction map changing with iteration}, our method did not detect cells well in the first iteration even though there were some cells in the original image. However, our method could detect those cells as the iteration progressed.

\begin{figure}[t]
    \centering
    \includegraphics[width=\linewidth]{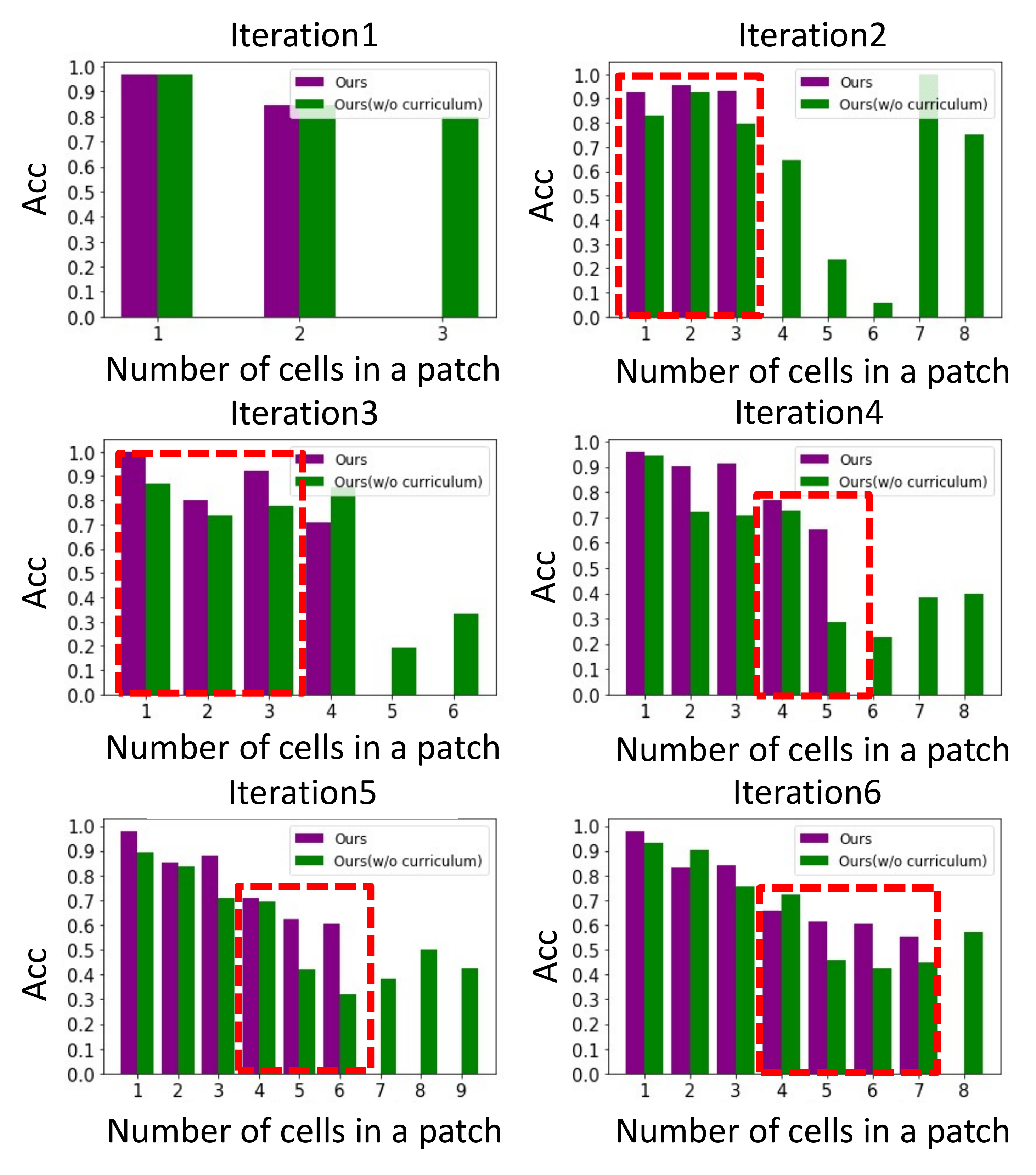}
    \caption{Comparison of the Bayesian discriminator performance with curriculum learning. The figure show the discriminator performance without curriculum learning (green bars) and with curriculum (purple bars). Red boxes indicate that the performance increases by adding the curriculum learning to the Bayesian discriminator.}
    \label{fig:perfomance on discriminator}
\end{figure} 

\begin{figure}[t]
    \centering
    \includegraphics[width=0.9\columnwidth]{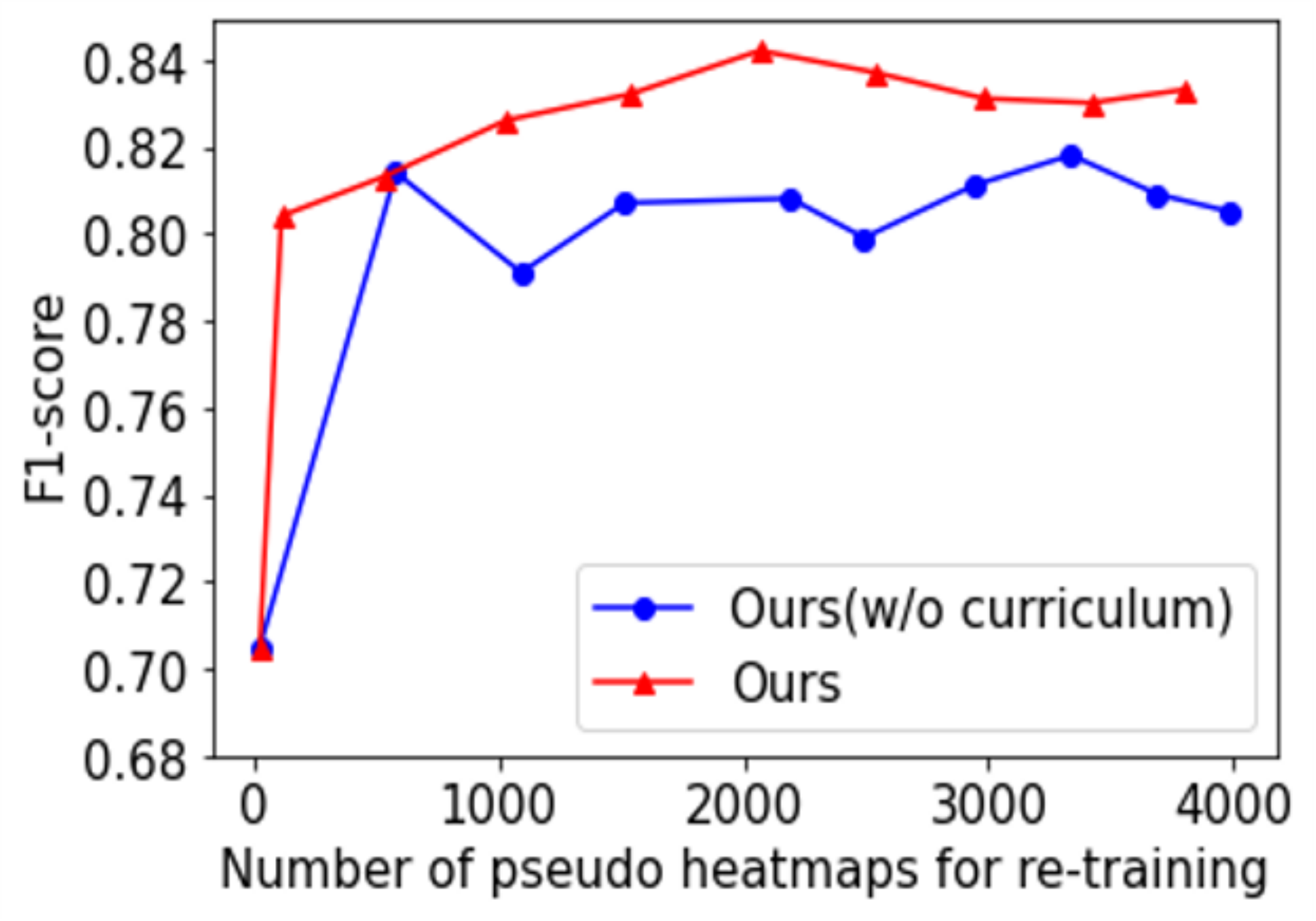}
    \caption{Detection performance of the proposed method with iteration and effect of curriculum learning. The red plot is our method with curriculum and the blue plot is our method without curriculum learning}
    \label{fig:the proposed method with iteration}
\end{figure}

\begin{table*}[t]
    \centering
    \caption{Effectiveness of the proposed method on different domains.}
    \label{tab3}
    \begin{tabular}{c|c|c|c|c|c}
    & Pseudo-Labeling & Discriminator & Bayesian dropout & Curriculum Learning & F-score\\\hline
    Baseline(\citeauthor{nishimura2019weakly}) &  & &  & &0.770\\
    Ours& \checkmark &  &  & &0.785\\
    Ours& \checkmark & \checkmark & & &0.831\\
    Ours(\citeauthor{cho2021cell})& \checkmark & \checkmark &\checkmark  & &0.855\\
    Ours& \checkmark & & & \checkmark &  0.796\\
    Ours& \checkmark & \checkmark& \checkmark &\checkmark & \textbf{0.866}
    \end{tabular}
\end{table*}

\subsection{Ablation study}
\textbf{Ablation study of each module in the proposed method:} Table \ref{tab3} shows the effectiveness of each module of our method in different domains. The F-score is the average of all the pairs of domains (14 pairs).
The 'Pseudo-Labeling' (pseudo-cell-position heatmap) method indicates that the method does not select pseudo-cell-position heatmaps and uses all of them on the target domain. 
'Discriminator' indicates that pseudo heatmaps were selected by referring to the output score of the CNN without using Bayesian dropout-based uncertainty. 'Bayesian Dropout' indicates that pseudo heatmaps were selected on the basis of uncertainty without curriculum learning.
These results confirm that using all of the pseudo-cell-position heatmaps (Pseudo-Labeling) in the target domains is not suitable for cell detection in different domains since they contain many noisy labels. 
The Bayesian discriminator enabled the proposed method to outperform the CNN discriminator that can not estimate uncertainty for pseudo labels.
In addition, the curriculum learning further improved detection performance.
This is because each module contributed to selecting positive heatmaps that are more likely to be correct. Additionally, we conducted the experiments to select pseudo labels through the curriculum learning without considering the uncertainty.  Interestingly, the detection performance improved from ours with pseudo-labeling  even if it was lower than that of using the uncertainty.

\textbf{Ablation study of the curriculum learning:} Fig. \ref{fig:perfomance on discriminator} shows the performance on the selected pseudo-cell-position heatmaps of our methods without and with curriculum learning at each iteration based on the number of cells in a patch. This experiment was performed under C2C12 (C→F) and the discriminator performance was calculated with the evaluation metric described in Section \ref{sec:Discriminator performance}. Note that we took the number of peaks in a patch to be the number of cells in a patch in this experiment since we did not know the true number in our problem setting. 
In iteration 1 (left figures), the accuracy on the images that contain 3 cells in ours (w/o curriculum) is worse than on those containing fewer cells. Ours with curriculum learning only selected the images that contain 1 or 2 cells (does not select 3 cells), and thus the selected pseudo heatmaps are more correct than ours (w/o curriculum).
Re-training using these better pseudo heatmaps improved the accuracy for pseudo label selection in the next step, as shown in the red box in Fig. \ref{fig:perfomance on discriminator}.
Accordingly, this good selection contributed to the later iteration.
This analysis confirms the effectiveness of the curriculum learning based on the number of cells. We consider that more accurate pseudo-cell-position heatmaps are important, even if the number of pseudo labels is smaller.

Fig. \ref{fig:the proposed method with iteration} shows the detection performance versus the number of pseudo heatmaps used for re-training the detection network for the experiment on C2C12 (C→F), where the red (blue) line indicates our method with (without) curriculum learning.
The performance of ours (w/o curriculum) increases the detection performance at the beginning of learning but the improvement saturates and does not obtain any further improvement by increasing the pseudo labels.
We consider that this is because incorrect pseudo heatmaps were selected in the early iteration of learning.
In contrast, ours with curriculum learning made further improvements as the iterations increased.
We consider that the selected pseudo-cell-position heatmaps by introducing curriculum learning are more correct than those without curriculum learning.

\begin{figure}[t]
    \centering
    \includegraphics[width=\columnwidth]{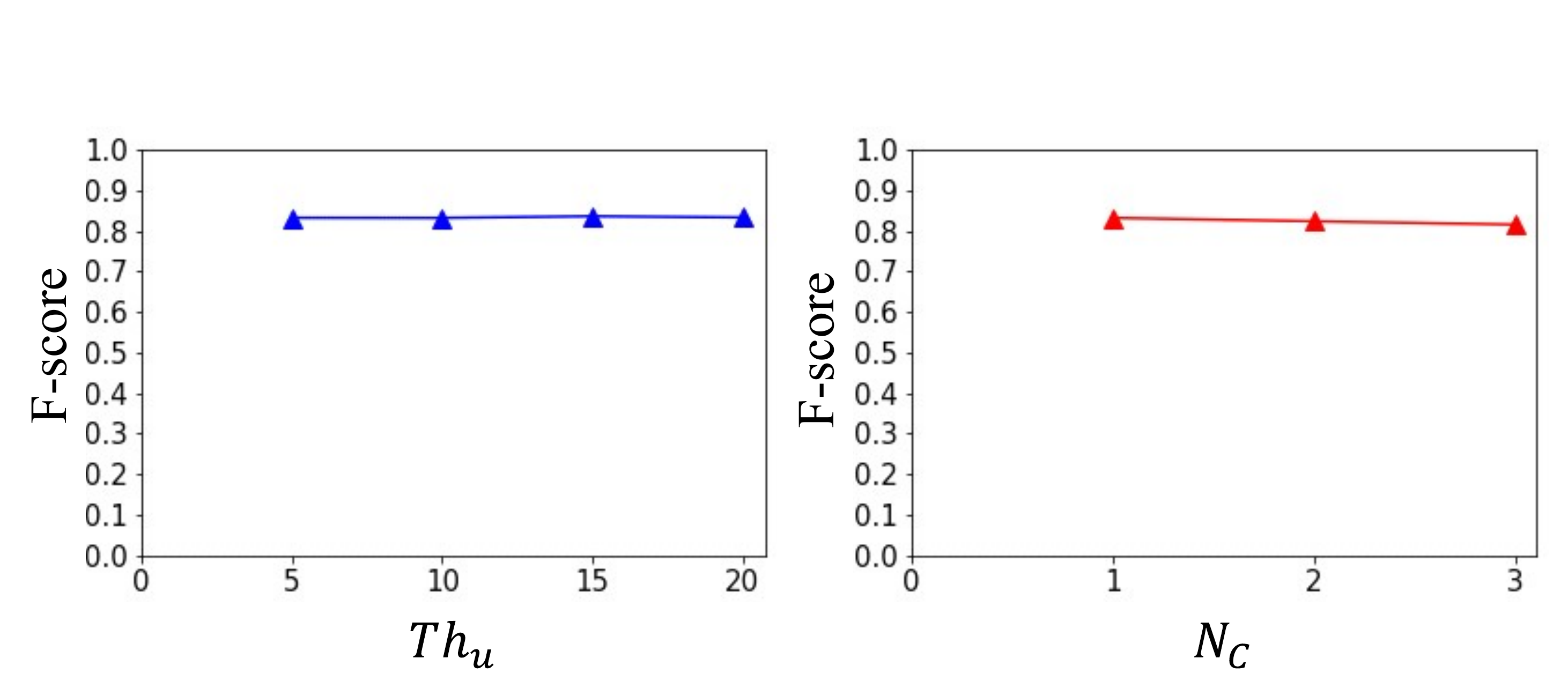}
    \caption{Parameter sensitivity analysis. {\bf Left:} the detection performance in terms of $th_u$. {\bf Left:} the detection performance in terms of $N_c$.}
    \label{fig:paramete_sensitivity}
\end{figure}

\begin{figure}[t]
    \centering
    \includegraphics[width=\columnwidth]{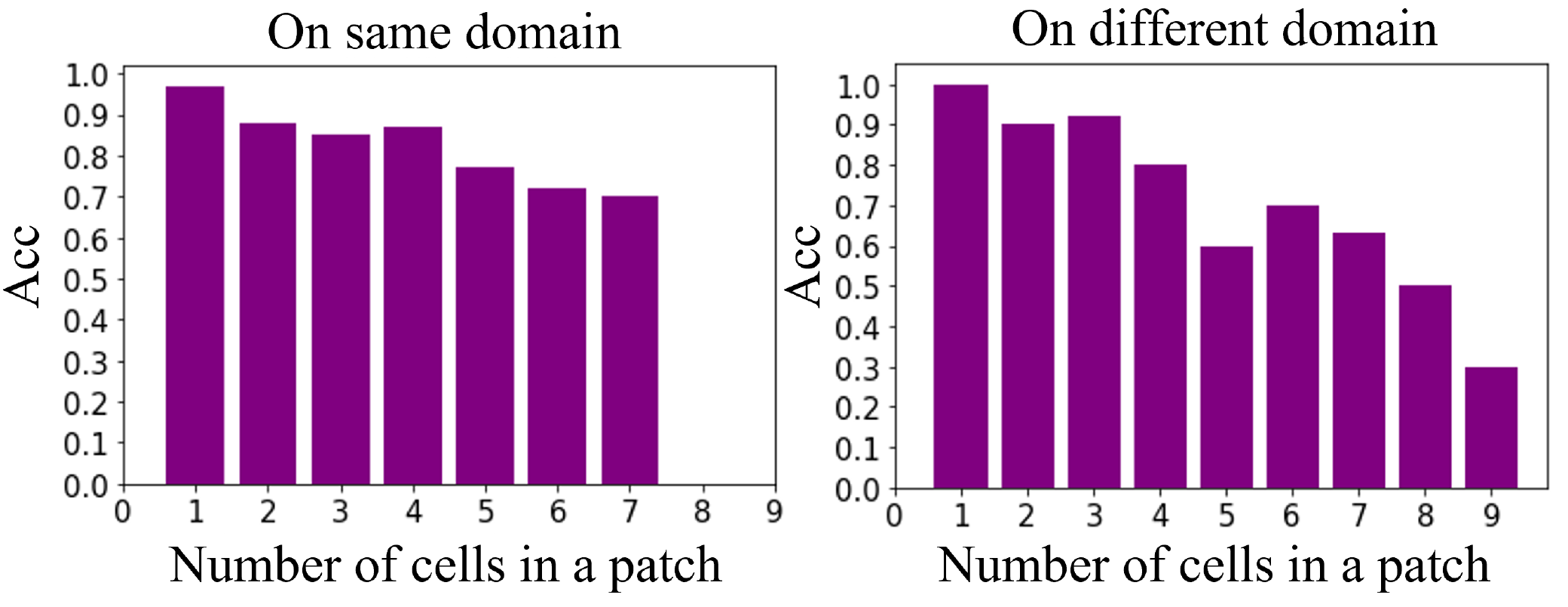}
    \caption{Performance of Bayesian discriminator.}
    \label{fig:bayesian on domain}
\end{figure}

Fig. \ref{fig:paramete_sensitivity} shows the hyper-parameter sensitivity of our method on $th_u$ and $N_c$. We varied $th_u$ from 5 to 20 (5, 10, 15, 20) and $N_c$ from 1 to 3 (1,2,3). The result shows the robustness of our method on $th_u$ and $N_c$.

Fig. \ref{fig:bayesian on domain} shows the performance of the Bayesian discriminator on the same domain (C→C) and the different domain (B→C). The Bayesian discriminator can select pseudo labels more correctly in the same domain than in the different domains. In the different domains, the accuracy of the discriminator decreases when the number of cells is large. Therefore, the curriculum learning is effective in the different domain setting. This result supports our conclusion that the pseudo labels selected by the Bayesian discriminator from the same domain were more accurate than those in different domains.

\section{Conclusion}
We proposed a domain adaptation method for cell detection based on semi-supervised learning by selecting pseudo-cell-position heatmaps using the uncertainty and the curriculum learning. The experiment results using various combinations of domains demonstrated the effectiveness of the proposed method, which improved the performance of detection on non-annotated cells on different domains. Also, the analysis of quantitative results on each iteration demonstrated that the method incrementally extended the domain from the source and target step by step.
Domain adaptation for the large domain shift such as different types of cells or microscopy, which are remained in general objects, is our future work. 

\section*{Acknowledgements}
This work was supported by JSPS KAKENHI Grant Number JP20H04211 and JP21K19829.

\bibliographystyle{model2-names.bst}\biboptions{authoryear}
\bibliography{refs}

\end{document}